\documentclass[preprint,12pt]{elsarticle}

\usepackage{amssymb}
\usepackage{amsmath}
\usepackage{booktabs}
\usepackage{algorithm}
\usepackage{algpseudocode}
\usepackage{multirow}
\usepackage{xcolor}
\usepackage{url}
\usepackage{subcaption}
\usepackage{threeparttable}
\usepackage{verbatim}

\algrenewcommand\algorithmicrequire{\textbf{Input:}}
\algrenewcommand\algorithmicensure{\textbf{Output:}}

\journal{Applied Soft Computing}

\begin{document}

\begin{frontmatter}



\title{Relational Multi-Agent Reinforcement Learning for Dynamic Pricing in High-Speed Railway Markets}


\author[affi1]{Enrique Adrian Villarrubia-Martin}
\ead{enrique.villarrubia@uclm.es}
\author[affi2]{David Muñoz-Valero\corref{cor1}}
\ead{david.munoz@uclm.es}
\author[affi1]{Luis Rodriguez-Benitez}
\ead{luis.rodriguez@uclm.es}
\author[affi3]{Giovanni Montana}
\ead{g.montana@warwick.ac.uk}
\author[affi1]{Luis Jimenez-Linares}
\ead{luis.jimenez@uclm.es}

%% Corresponding author
\cortext[cor1]{Corresponding author}

\affiliation[affi1]{organization={Department of Technologies and Information Systems, Universidad de Castilla-La Mancha}, %Department and Organization
            addressline={Paseo de la Universidad 4}, 
            city={Ciudad Real},
            postcode={13071}, 
            %state={},
            country={Spain}}

\affiliation[affi2]{organization={Department of Technologies and Information Systems, Universidad de Castilla-La Mancha}, %Department and Organization
            addressline={Avenida Carlos III, s/n}, 
            city={Toledo},
            postcode={45071}, 
            %state={},
            country={Spain}}

\affiliation[affi3]{organization={Warwick Manufacturing Group, University of Warwick}, %Department and Organization
            addressline={Gibbet Hill Road}, 
            city={Coventry},
            postcode={CV4 7AL}, 
            %state={},
            country={UK}}

%% Abstract
\begin{abstract}
In liberalised railway systems, operators must set prices dynamically in an environment with partial observability, as they retain private information about their objectives and performance, where regulatory constraints prohibit communication or direct information exchange between competitors to prevent explicit collusion. Consequently, agents must learn to infer strategic interactions only from observable market data which presents a significant challenge for multi-agent reinforcement learning, where standard approaches typically treat observations as unstructured vectors, ignoring the underlying market topology that governs strategic interactions. To address this, an entity graph modelling approach is proposed, which represents the environment as a graph of operational units, rather than decision-making agents or static infrastructure, encoding competition, coordination, and connectivity relations between entities. Then, an extension of the multi-agent twin delayed deep deterministic policy gradient algorithm with graph-based representation learning processes the features of the entities through a multi-layer relational graph convolutional network and aggregates them via a learnt attention mechanism. Experimental results in a rail pricing reinforcement learning environment show that this novel framework achieves higher revenue and stability in two different settings of increasing market complexity compared to a representative selection of relational and non-relational baselines. The code is publicly available at: \url{https://github.com/Kinrre/RelationalRailPricing-RL}.
\end{abstract}


%\item An entity graph MARL approach to capture fine-grained strategic dependencies.
%\item Entities are connected through competition, coordination, and connectivity relations.
%\item MATD3 algorithm is extended with graph representation learning and attention pooling.
%\item Higher revenue than relational and non-relational baselines in railway pricing.

%% Keywords

\end{frontmatter}

\section{Introduction}\label{sec1}

The Multi-Agent Reinforcement Learning (MARL) community has increasingly adopted graph-based approaches to model complex environments \cite{Jiang2020GRAPHLEARNING,goeckner2024graph}. Graphs are a common and very powerful practice to naturally and efficiently capture the structural relations that are fundamental to multi-agent interaction and reasoning \cite{utke2025investigating}. The environments are characterised by a highly dynamic nature, where agents are constantly moving and their neighbourhood relations evolve rapidly, which creates a need for architectures capable of capturing abstract relational representations. Graph Neural Networks (GNNs) are framed as the ideal solution, precisely because they operate natively on graph structures, allowing the learning process to automatically adapt to the changing topology of the environment \cite{Ardjmand2026}. As agents modify their connections, GNNs maintain the ability to model interactions through message passing operations, which transforms relational variability from a limitation into an advantage, without the need to adapt the architecture of the model.

Focusing on railway systems, the field of application selected for experimentation, graph-based representations have emerged as the standard approach for modelling complex transportation networks. Station-based graphs, where vertices encode physical stations and edges denote direct rail connections, remain the predominant paradigm in recent railway optimisation literature \cite{Liu2025NetworkLearning}. This approach is widely used for network topology analysis, vulnerability assessment, and capacity planning \cite{Li2024}. However, these models, where stations are encoded as nodes, are typically not designed to capture operational aspects, such as service differentiation, multi-operator dynamics, and market-level interactions in liberalised high-speed passenger markets \cite{Gutirrez-Hita2022}. To bridge this gap, this work introduces an entity graph-based modelling approach, where operational entities constitute the vertices, and competitive, coordination, and connectivity relations are encoded as heterogeneous edges. With these relations encoded in the graph structure, GNN layers propagate messages across the network while attention mechanisms can selectively aggregate the most relevant features for each agent. This produces learnable relational representations that can be used to support strategic decision-making in multi-agent environments.

Based on the entity graph-based modelling framework \cite{Agarwal2020,nayak2023scalable}, the proposed model learns a relational state representation for each agent through a three-stage process. First, it individually encodes the characteristics of all entities in the environment. Then, it enriches them by allowing entities to exchange contextual information with their neighbours through the network of relations. Finally, it applies an attention mechanism that allows each agent to selectively combine the most relevant information from all entities, thus generating a compact and personalised state of the environment. These relational representations are subsequently integrated into an actor-critic architecture augmented with graphs that operates under the Centralised Training with Decentralised Execution (CTDE) \cite{Foerster2018CounterfactualGradients,Lowe2017Multi-agentEnvironments} paradigm. This architecture allows agents to access global information during training, when graph-based embeddings are constructed and refined, while maintaining decentralised policies that preserve privacy at runtime. The proposed method extends the Multi-Agent Twin Delayed Deep Deterministic Policy Gradient (MATD3) \cite{ackermann2019reducing} algorithm with graph representation learning capabilities, capturing the relational structure of the environment and addressing the challenge of multi-agent strategic interaction under partial observability with information asymmetry, which is typical of competitive market environments such as the liberalised passenger rail transport market.

\subsection{Contributions}

The contributions of this work can be summarised as follows:

\begin{itemize}
    \item An entity graph modelling approach for MARL that encodes the environment as a graph of operational units rather than decision-making agents or static infrastructure. This approach models competition, coordination, and connectivity relations as heterogeneous edges between entities, providing a relational inductive bias \cite{Zambaldi2019DeepBiases} to capture fine-grained strategic dependencies in partially observable environments.
    
    \item An extension of the MATD3 algorithm with graph-based representation learning. The proposed method specifically incorporates a relational state representation module that processes heterogeneous entity features through a multi-layer Relational Graph Convolutional Network (R-GCN) and aggregates them via a learnt attention mechanism to learn policies that dynamically prioritise strategically relevant services. In addition, it incorporates a gradient-stopping operation to decouple representation learning from policy optimisation, which, although it can come at the cost of a reduced total revenue, stabilises the training dynamics of each agent individually.
    
    \item An experimental study of the framework in a railway pricing reinforcement learning environment \cite{Villarrubia-Martin2025DynamicLearning} in two different scenarios, with increased revenue and stability in both settings. The performance of the method is compared against a representative selection of relational and non-relational MARL baselines. Several ablation studies have been conducted to measure the relative contribution of each edge type, the impact of the network depth on agent performance, the importance of the learnt attention mechanism, and the effect of decoupling the representation learning from the actor. Furthermore, the learnt graph embeddings are also explored.
\end{itemize}

The rest of this work is structured as follows: First, Section \ref{sec:background} reviews related work. Then, Section \ref{sec:preliminaries} introduces the formal framework of partially observable Markov Game (MG). Then, Section \ref{sec:methods} presents the entity-graph relational actor-critic framework. After that, Section \ref{sec:experiments} evaluates the effectiveness of the framework, and finally, Section \ref{sec:conclusions} outlines the conclusions and future research directions.

% --------------------
\section{Related work}\label{sec:background}

First, Section \ref{sec:backMARLDynamicPricing} reviews MARL in the context of dynamic pricing. Secondly, Section \ref{sec:backGNN} examines how graph-based representations can capture the relational structure inherent in multi-agent environments. Lastly, Section \ref{sec:backRailway} discusses the application of Deep Reinforcement Learning (DRL) to railway systems.

\subsection{Multi-agent reinforcement learning for dynamic pricing} \label{sec:backMARLDynamicPricing}

MARL extends the single-agent reinforcement learning framework to scenarios in which multiple agents interact in a shared environment to achieve individual or collective goals \cite{marl-book}. In contrast to single-agent settings, MARL environments are inherently non-stationary, as each agent's policy evolves during training, thus altering the environment for other agents. This interaction is formally modelled through game theory, and settings involving both competitive and cooperative objectives are referred to as coopetition \cite{Pang2025Data-drivenNetworks}. A particularly influential paradigm for addressing non-stationarity that maintains scalable, decentralised execution is CTDE \cite{Foerster2018CounterfactualGradients,Lowe2017Multi-agentEnvironments}. Under CTDE, agents share global information during training to stabilise learning, while retaining independent policies at execution time. Within this paradigm, MADDPG \cite{Lowe2017Multi-agentEnvironments} adapts the deterministic policy gradient approach for continuous action spaces in mixed cooperative-competitive environments, Multi-Agent Proximal Policy Optimisation (MAPPO) \cite{Yu2022TheGames} extends proximal policy optimisation to cooperative settings, and Multi-Actor Attention Critic (MAAC) \cite{Iqbal2019Actor-attention-criticLearning} incorporates attention mechanisms to dynamically weight the relevance of other agents when computing centralised Q-values.

Dynamic pricing, also known as pricing intelligence, involves adjusting prices in real time based on supply and demand to maximise profits and align with market conditions. Within this context, for electric vehicle charging, \cite{Hou2023} proposed a MARL mechanism design framework to simultaneously determine optimal charging prices across multiple stations, modelling station–user interaction as a mechanism design problem and station–station cooperation as an MG solved via MADDPG. Similarly, for hydrogen fuel cell vehicle refuelling, \cite{Fraija2024DeepConstraints} employed a Multi-Agent System (MAS) to coordinate refuelling schedules and determine prices, improving demand satisfaction and traffic flow in microgrids. Shifting to other domains, in telecommunications, \cite{Sun2023CompetitiveApproach} framed the interaction between mobile virtual network operators and users as a Stackelberg game, solving it using a Multi-Agent Deep Q-Network (MADQN). Moreover, in smart grids, \cite{Ma2024OptimalAlgorithm} proposed a distributed multi-agent optimisation approach for resolving supply-demand imbalances while strengthening privacy and autonomy.

Despite these advancements, dynamic pricing in high-speed railways introduces complex relational dependencies, such as cooperative dynamics with multi-operator services and overlapping market competition. Furthermore, the partial observability and the prohibition of communication between agents in this domain demand that agents infer strategic interactions only from observable market data. This motivates the use of graph-based representations, where interacting entities and their relationships are naturally encoded as nodes and edges in a graph structure.

\subsection{Graph neural networks in multi-agent reinforcement learning} \label{sec:backGNN}

Multi-agent environments frequently involve relational information, where agents interact with each other through dependencies such as competition or coordination. However, traditional deep MARL approaches encode a state as a flat vector, a representation that discards the structural relations between agents, forcing neural networks to implicitly discover patterns that could be encoded explicitly. GNNs can be used as an alternative that preserves this relational structure \cite{Kipf2017Semi-supervisedNetworks}. By modelling the environment as a graph, where vertices denote decision-making entities and edges encode their relationships, agents are able to reason explicitly about their interdependencies. For instance, \cite{Jiang2020GRAPHLEARNING} introduced graph convolutional reinforcement learning, which constructs a graph where each agent corresponds to a vertex and edges connect agents that influence each other's outcomes. Rather than using fixed rules to determine these connections, the method employs multi-head attention mechanisms as relation kernels, learning how agents interact with their neighbours, with temporal relation regularisation further encouraging cooperative behaviours over time. Similarly, \cite{Liu2020Multi-AgentNetwork} developed G2ANet, which introduces a two-stage attention network combining hard attention, to identify and eliminate irrelevant agent connections, and soft attention, to weigh the importance of remaining interactions.

Other approaches have explored richer graph structures that incorporate additional context from the environment. In \cite{nayak2023scalable}, they proposed InforMARL, which constructs an agent-entity graph connecting agents not only to other agents but also to surrounding entities such as obstacles and goals. At each time step, the framework builds a graph linking each agent to nearby entities within its local neighbourhood, employing a GNN with multi-head attention to aggregate this information into fixed-size representations for actor-critic networks. This local aggregation is more sample-efficient than concatenating global information, as it allows agents to focus on locally relevant context while maintaining a scalable representation. Additionally, \cite{goeckner2024graph} developed a GNN-based MARL method for resilient distributed coordination of multi-robot systems operating under challenging real-world conditions. The approach employs MAPPO using a Multilayer Perceptron (MLP) critic for training and a GNN actor that incorporates both node and edge features through a modified GraphSAGE \cite{hamilton2017inductive} architecture with multi-layer message passing. A key innovation is a neighbour scoring mechanism which evaluates and selects the graph edges that agents should traverse for discrete wayfinding.

Several successful applications can be found across multiple domains. In traffic signal control, \cite{Yoon2021TransferableRepresentation} introduced a graph state representation, and in a similar way, \cite{yang2023hierarchical} proposed the HG-M2I algorithm, which combines a hierarchical graph representation learning module with a multi-agent mutual information framework. In a related line of work on traffic management, \cite{Liu2025NetworkLearning} applied DRL with graphs for the extension planning of metro networks, improving the resilience of urban critical infrastructures. Beyond transportation, \cite{Pang2026} proposed Multi-Relational Graph Reinforcement Learning (MRGRL) for dynamic flexible job-shop scheduling, explicitly modelling multi-type dependencies and competition patterns through heterogeneous graphs, demonstrating strong generalisation under unstable manufacturing conditions.

Although these graph-based approaches are effective for reasoning about relational structure, they predominantly model agents themselves as graph nodes or static physical infrastructure, which limits their ability to capture the fine-grained operational dependencies that drive strategic interactions, as is the case of dynamic pricing in liberalised railway systems.

\subsection{Deep reinforcement learning for railway systems} \label{sec:backRailway}

The application of DRL to railway systems has grown considerably in recent years, with a potential for optimising operations and improving efficiency. In the context of traffic management, DRL has been applied to real-time trajectory generation for trains to ensure punctuality and energy efficiency \cite{Ning2022DeepOptimization}, as well as to Train Timetable Rescheduling (TTR) and the Vehicle Rescheduling Problem (VRSP), where agents must restore normal operations rapidly following disruptions \cite{Yue2024ReinforcementRepresentation, Mohanty2020Flatland-rl:Trains}. Further applications include predictive maintenance scheduling \cite{Mohammadi2022APlanning, Arcieri2024POMDPMaintenance}, autonomous train control \cite{Feng2021ANetwork, Cui2020Knowledge-basedRailway}, and interval control under safety constraints using constrained Markov decision processes \cite{Lin2024TrackingLearning}. Simulation environments such as Flatland-RL \cite{Mohanty2020Flatland-rl:Trains} have played an important role in supporting the safe development and benchmarking of these approaches \cite{Li2021ScalableChallenge}.

However, existing DRL applications in railways are predominantly concerned with operational problems rather than market-driven strategies. Dynamic pricing introduces additional challenges where agents must model the distinct passenger behaviours, account for interdependencies arising from multi-operator itineraries, as well as the inherent tension between competition and cooperation in a partially observable market. To address this gap, \cite{Villarrubia-Martin2025DynamicLearning} introduced RailPricing-RL, a parameterisable MARL environment designed specifically for dynamic pricing in high-speed railway markets where operators must balance competition and cooperation with connecting services, incorporating microscopic passenger decision-making via Random Utility Models (RUM). The current study extends this approach by introducing a novel relational state representation that explicitly captures these structural dependencies.

\section{Preliminaries} \label{sec:preliminaries}

\begin{figure}[!ht]
    \centering
    \includegraphics[width=0.77\textwidth]{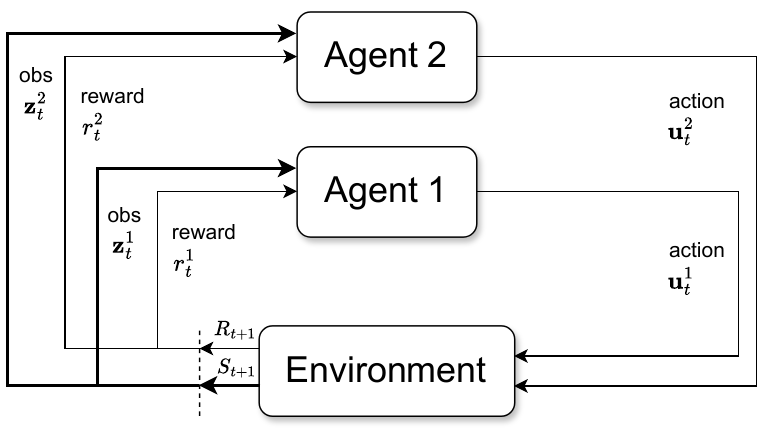}
    \caption{MARL interaction in a partially observable MG.}
    \label{fig:preliminaries}
\end{figure}

This section introduces a partially observable MG \cite{Littman1994MarkovLearning}, a stochastic game that satisfies the Markov property, with mixed incentives in which agents have both aligned and competing objectives. In an MG, a multi-agent extension of an MDP, multiple agents act as independent decision-makers, each with their own set of actions and reward functions, as shown in Figure \ref{fig:preliminaries}.

A key property is that both state transitions and individual rewards depend on the joint actions of all agents, creating a complex interdependence of strategies and outcomes in which no single agent fully controls the environment dynamics. Formally, it is defined by the tuple:

\begin{equation*}
M = \langle S, U, P, r, Z, O, A, \gamma \rangle
\end{equation*}

where $S$ is the set of possible states, with $s \in S$ denoting the current state of the environment. The set $A = \{1, \dots, n\}$ represents the $n$ agents, where each agent $a \in A$ selects actions $u^a \in U$ at each time step. These actions collectively form a joint action $\mathbf{u} \in \mathbf{U} \equiv U^n$, which induces a transition to a new state $s'$ according to the state transition function $P(s' \mid s, \mathbf{u}) : S \times \mathbf{U} \times S \to [0, 1]$.

In contrast to Decentralised Partially Observable Markov Decision Processes (DPOMDP) \cite{Oliehoek2016APOMDPs}, typically used in fully cooperative settings where all agents share a common reward function, stochastic games allow competitive, cooperative, or mixed forms of interaction, the latter often referred to as coopetition \cite{Pang2025Data-drivenNetworks}. Consequently, each agent is assigned an individual reward function, defined as $r^a(s, \mathbf{u}) : S \times \mathbf{U} \to \mathbb{R}$, which depends on the global state and the joint action. Partial observability is intrinsic to realistic multi-agent settings, contrasting with perfect information games, where the full state is observable. This is the case in railway market environments, where operators retain private information about their objectives, behaviour, performance, etc. Formally, each agent receives an observation $\mathbf{z}^a \in Z^a$, which provides partial information about the global state. It is determined by the agent-specific observation function $O(s, a) : S \times A \to Z^a$, where $Z^a$ denotes the observation space for agent $a$. Furthermore, agents operate under strict information asymmetry without explicit communication, where each agent $a$ can observe only its own reward signal $r^a$ and local observation $\mathbf{z}^a$, whereas the rewards, policies, and value functions of other agents remain hidden.

While traditional approaches treat the observations as unstructured vectors, real-world MAS often exhibit a rich relational structure with entities connected by specific relations that influence strategic interactions. To capture the structural relations inherent in multi-agent environments, the state is augmented with a graph $\mathcal{G}_t = (V, E, X)$, where $V$ denotes the set of vertices, $E \subseteq V \times V$ are the edges encoding relations between vertices, and $X = \{x_v\}_{v \in V}$ comprises node features. The graph evolves with the environment state at each time step $t$, yielding $\mathcal{G}_t = \mathcal{G}(s_t)$, and provides a structured representation that enables reasoning about relational dependencies beyond flat feature vectors. This is a separate augmentation of the state rather than an additional element of the tuple $M$. Finally, the goal of each agent $a$ is to learn a stochastic policy $\pi^a(u^a \mid \mathbf{z}^a) : Z^a \times U \to [0, 1]$, which maps observations to a probability distribution over actions. The collection of policies for all agents, $\boldsymbol{\pi} = (\pi^1, \dots, \pi^n)$, forms the joint policy. Thus, the objective of each agent is to maximise its expected discounted return, defined as:

\begin{equation*}
    J(\pi^a) = \mathbb{E}_{\mathbf{u} \sim \boldsymbol{\pi}, s \sim P} \left[ \sum_{t=0}^\infty \gamma^t r_t^a(s, \mathbf{u}_t) \right],
\end{equation*}

where $\gamma \in [0, 1]$ is the discount factor that determines the relative importance of immediate versus long-term rewards.

\section{Methods} \label{sec:methods}

This section proposes an entity graph-based relational actor-critic framework for MARL in environments with rich relational structure. First, Section~\ref{sec:entity_graph} introduces the entity-based graph modelling approach, then Section~\ref{sec:graph_representation} details the semantic embedding architecture, and finally Section~\ref{sec:marl_framework} presents the graph-augmented actor-critic architecture that integrates these representations.

\subsection{Entity-based graph modelling} \label{sec:entity_graph}

The first goal is to encode the environment as a heterogeneous graph to provide a relational inductive bias \cite{Zambaldi2019DeepBiases,Battaglia2018RelationalNetworks,Ugadiarov2025RelationalActor-Critic} overcoming the limitations of unstructured state representations common in MARL \cite{Yu2022TheGames}. Typically, nodes in these graphs are the agents themselves \cite{Jiang2020GRAPHLEARNING,Liu2020Multi-AgentNetwork} or the static physical infrastructure \cite{Liu2025NetworkLearning,Yoon2021TransferableRepresentation}. As a novel contribution, the entity graph, instead of focusing on decision-making agents such as railway companies, focuses on the operational nodes they control, referred to as entities. This paradigm shift moves beyond modelling decision-makers or passive infrastructure, towards encoding the strategic units of control themselves. The approach treats entities as the primary vertices in the graph, modelling the different relations between operational units within the MAS. It captures how strategic decisions affect and are affected by the broader ecosystem of interactions, rather than forcing neural networks to implicitly discover these patterns from flat feature vectors.

These entities in the railway domain take the form of specific train services, which constitute the resources and functional units on which agents operate. A service is defined as a scheduled journey between stations at a specific time with allocated capacity. The value of this entity approach lies in the explicit modelling of the relational dynamics that arise between these entities. Thus, competitive dependencies are captured, where services from different agents compete for the same market; coordination relations, that forms the agent's internal service portfolio; and connectivity dependencies, where one entity requires another to operate, so that passengers can transfer between services to form multi-operator routes. As a consequence, the framework can reason about how pricing decisions propagate through the service network.

To formalise this, at each time step $t$, the railway market is modelled as a directed graph $\mathcal{G}_t = (V, E, X)$ where vertices correspond to individual railway services and edges encode these market relations. Here, each vertex $v \in V$ corresponds to a unique service operated within the railway network by an agent $agent(v) \in A$. A service is a scheduled, multi-stop journey along a line of consecutive stations and serves a set of origin–destination markets $W(v)$, where each market $(w_o, w_d)$ is specified by its origin station $w_o$ and destination station $w_d$, defining an origin-destination pair where passengers seek transportation. The edge set is formed through three specific relation types described as follows, where $v_i,v_j$ are any two distinct vertices in $V$:

\begin{itemize}
    \item $E_{\mathrm{comp}} = \{(v_i, v_j) : W(v_i) \cap W(v_j) \neq \emptyset, \; date(v_i) = date(v_j), \; agent(v_i) \neq agent(v_j)\}$ denotes competition edges connecting services scheduled on the same date that operate in at least one common market $w$ but are managed by different agents.
    \item $E_{\mathrm{coord}} = \{(v_i, v_j) : date(v_i) = date(v_j), \; agent(v_i) = agent(v_j)\}$ represents coordination edges that link services operated by the same agent on the same date.
    \item $E_{\mathrm{conn}} = \{(v_i, v_j) : W_d(v_i) \cap W_o(v_j) \neq \emptyset, \; date(v_i) = date(v_j)\}$ defines connectivity edges as potential passenger connections between services scheduled on the same date, where $W_o(v) = \{w_o : (w_o, w_d) \in W(v)\}$ and $W_d(v) = \{w_d : (w_o, w_d) \in W(v)\}$ are the sets of origin and destination stations of the markets served by $v$.
\end{itemize}

Competition and coordination edges are symmetric, so the reverse edge $(v_j, v_i)$ is also added, whereas the connectivity relation is naturally directed from the arriving service to the departing service. Thus, the edge set $E$ is:

\begin{equation*}
    E = E_{\mathrm{comp}} \cup E_{\mathrm{coord}} \cup E_{\mathrm{conn}}.
\end{equation*}

\begin{figure}[!ht]
    \centering
    \includegraphics[width=0.58\textwidth]{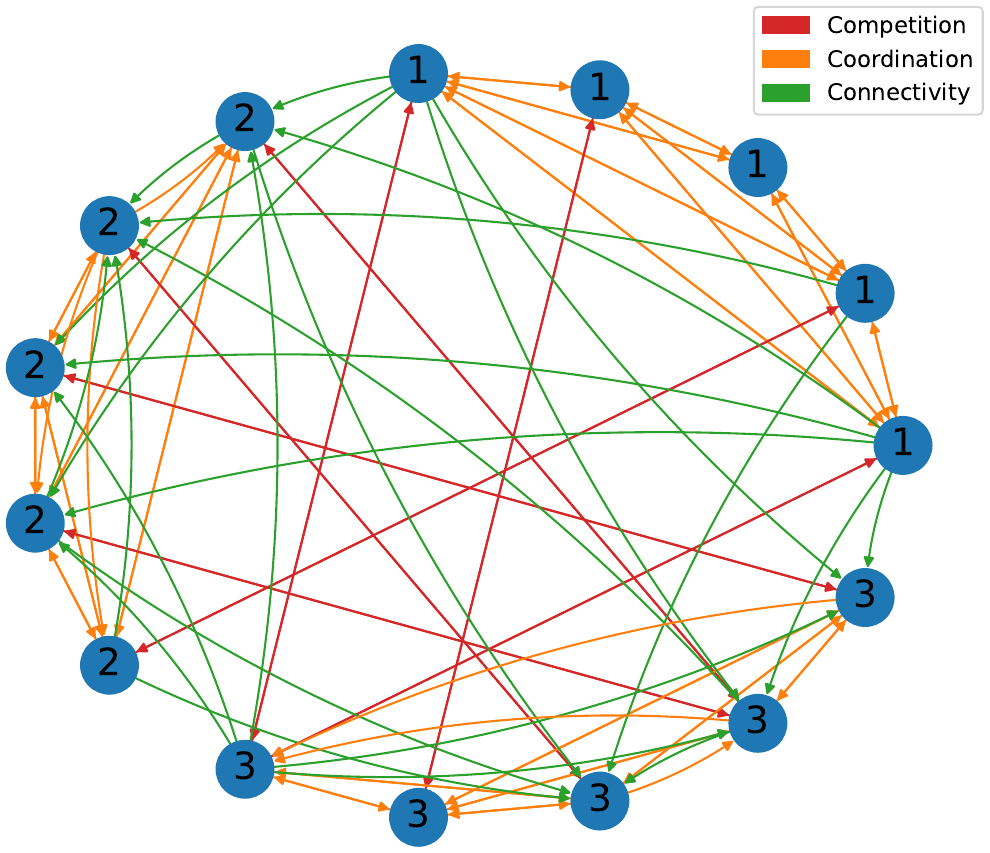}
    \caption{Entity graph representation for a railway network with 15 services scheduled on the same date by three agents. Red edges indicate competition, services in the same market by different operators, orange edges denote coordination, services operated by the same agent, and green edges represent connectivity, potential transfers between services.}
    \label{fig:service_graph}
\end{figure}

Figure~\ref{fig:service_graph} illustrates a concrete example of this entity graph, visualising the dense network of relations for a market comprising 15 services operated by three competing agents, where each node corresponds to a service, labelled by its controlling agent. Notably, the coordination edges densely interconnect within each agent's portfolio, competition edges link services in overlapping markets, and connectivity edges weave between services to form potential multi-operator routes.

\subsection{Relational state representation learning} \label{sec:graph_representation}

\begin{figure}[!ht]
    \centering
    \includegraphics[width=0.8\textwidth]{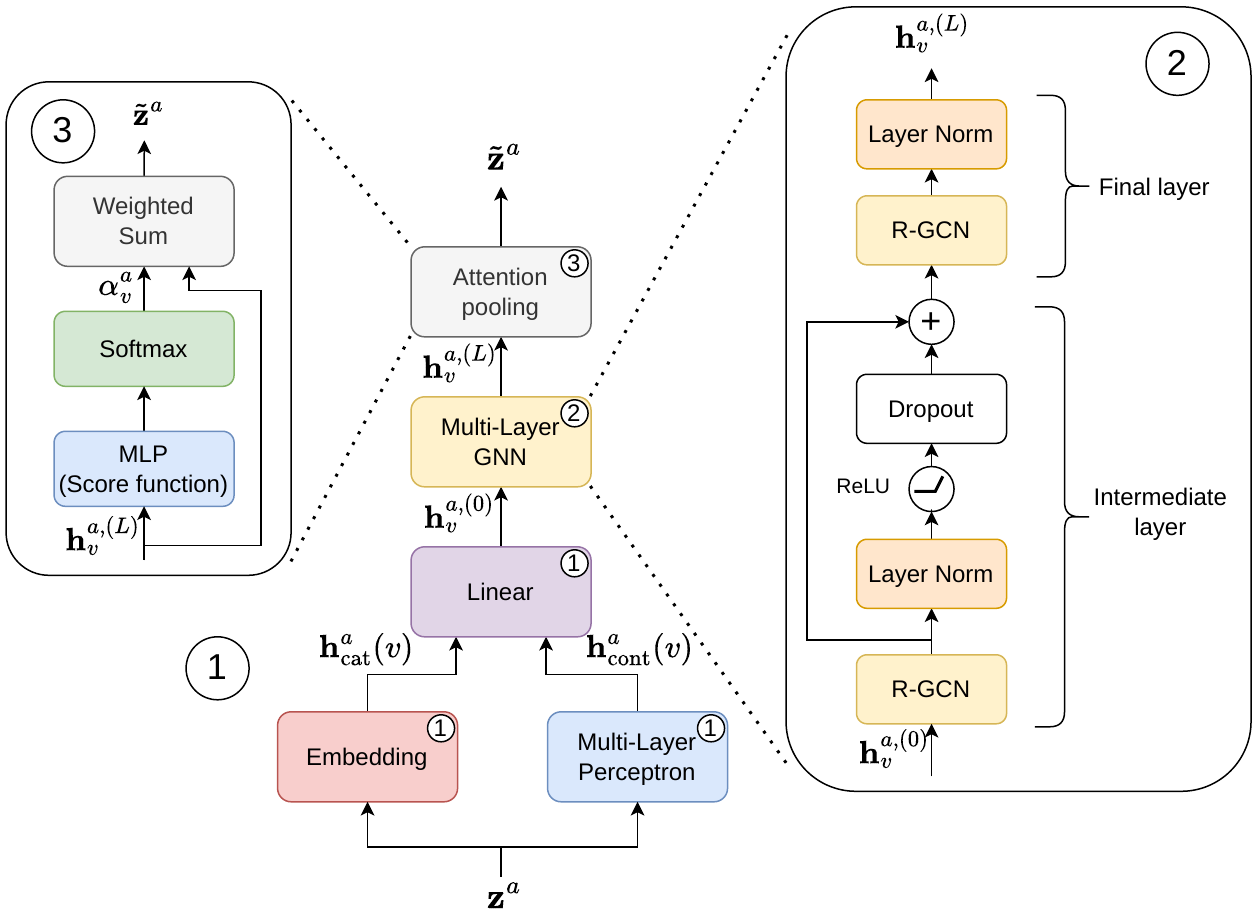}
    \caption{Architecture of the relational state representation module. For each agent, the entity features are first encoded, with categorical attributes through embeddings and continuous ones through an MLP, and then concatenated and projected by a linear transformation. A multi-layer R-GCN refines them by using message passing over the entity graph. Finally, a learnt attention pooling aggregates them into a fixed-size state vector.}
    \label{fig:relational_state_representation}
\end{figure}

This framework learns a relational state representation for each agent through a three-stage process (see Figure \ref{fig:relational_state_representation}): heterogeneous entity feature encoding, which processes categorical and continuous entity attributes through separate pathways (Section \ref{sect:heterogeneus}), relational context propagation, which performs message passing to create context-aware entity embeddings (Section \ref{sect:multilayer}), and agent-level state aggregation, which produces a fixed-size state vector for each agent, computed from the set of entity embeddings (Section \ref{sect:aggregation}).

\subsubsection{Encoding of heterogeneous entity features} \label{sect:heterogeneus}

The initial stage transforms the raw, heterogeneous features of each entity node into a unified dense representation, which serves as the input to an individual GNN. For each agent $a$, this transformation is achieved through parallel pathways designed to handle the distinct properties of categorical and continuous data before fusing them into a common latent space. Each entity $v$ is characterised by a heterogeneous feature vector comprising categorical attributes, such as origin and destination stations, seat class, operator, line, corridor, time slot and rolling stock type, and continuous market indicators used in dynamic railway pricing models. Each categorical attribute type $k$ is transformed through a learnable embedding function:

\begin{equation*}
\mathcal{E}_k^a: \mathbb{N} \to \mathbb{R}^{d_e}
\end{equation*}

which maps discrete category indices to dense embedding vectors of dimension $d_e$. This enables learning continuous semantic representations of the respective categorical attributes, which are subsequently concatenated to form a unified categorical feature vector:

\begin{equation*}
    \mathbf{h}_{\mathrm{cat}}^a(v) = \underset{k \in \mathcal{K}}{\parallel} \mathcal{E}_k^a(v_k)
\end{equation*}

where $\mathcal{K}$ is the set of all categorical attributes for entity $v$, and $\parallel$ denotes the concatenation operation.

The dynamic market data associated with service $v$, comprising prices $p_{s,c}$ and sales volume $n_{w,c}$ for each service-seat combination and the total service profit $r_{w,c}$ are flattened into a single vector $\mathbf{x}_{\mathrm{cont}}^a(v)$. To enforce partial observability, agent $a$ observes complete information for its own services, but for competitor services the private ticket-sales information, that is, sales volume per service-seat and total service profit, is masked with zeros. Although zero sales is a valid observation, it cannot be confused with a masked one, as masking is applied only to competitor services which can be identified by the operator feature. This vector is then processed by an MLP to capture complex, nonlinear interactions and dependencies within the market signals, such as price elasticity effects:

\begin{equation*}
    \mathbf{h}_{\mathrm{cont}}^a(v) = \mathrm{MLP}^a(\mathbf{x}_{\mathrm{cont}}^a(v))
\end{equation*}

Finally, the categorical and continuous vectors are concatenated and projected into the target hidden dimension $d_h$ by a linear transformation:

\begin{equation*}
    \mathbf{h}_v^{a,(0)} = \mathbf{W}_{\mathrm{proj}}^a[\mathbf{h}_{\mathrm{cat}}^a(v) \parallel \mathbf{h}_{\mathrm{cont}}^a(v)] + \mathbf{b}_{\mathrm{proj}}^a.
\end{equation*}

This projection yields the initial node embedding $\mathbf{h}_v^{a,(0)} \in \mathbb{R}^{d_h}$ for each entity, a unified embedding that captures the intrinsic properties independently of the graph structure, without incorporating any relational information from neighbouring entities. These entity-level representations then serve as input for the relational context propagation stage, where message passing over the graph enriches them with information from the entity's neighbourhood.

\subsubsection{Relational context propagation through a multi-layer GNN} \label{sect:multilayer}

The initial entity embeddings $\mathbf{h}_v^{a,(0)}$ capture intrinsic properties but lack awareness of the broader market structure. To incorporate this relational context, a multi-layer GNN that performs message passing across the heterogeneous entity graph, is used. Specifically, a stack of R-GCN \cite{schlichtkrull2018modeling} layers is employed to iteratively aggregate information from neighbouring entities. At each layer $\ell \in \{1, 2, \ldots, L\}$, the hidden representation $\mathbf{h}_v^{a,(\ell)}$ of entity $v$ is updated through message passing over the graph structure $\mathcal{G}_t$:

\begin{equation*}
    \mathbf{h}_v^{a,(\ell)} = f^{(\ell)}\left(\sum_{r \in \mathcal{R}} \sum_{u \in \mathcal{N}_r(v)} \frac{1}{|\mathcal{N}_r(v)|} W_r^{(\ell)} \mathbf{h}_u^{a,(\ell-1)} + W_0^{(\ell)} \mathbf{h}_v^{a,(\ell-1)} +\mathbf{b}^{(\ell)}\right), \ v\in V
\end{equation*}

where $\mathcal{N}_r(v)$ denotes the neighbourhood of node $v$ under relation $r$ connected through competition, coordination, or connectivity edges. Each relation weight matrix $W_r^{(\ell)}$ transforms the aggregated neighbour representations, whereas $W_0^{(\ell)}$ corresponds to a self-loop that introduces a special relation type to each node, to ensure that the node at layer $\ell$ is also informed by the corresponding representation at the previous layer $\ell-1$. In addition, $\mathbf{b}^{(\ell)}$ is a learnable bias vector and the activation function $f^{(\ell)}$ introduces nonlinearity. The normalisation constant $|\mathcal{N}_r(v)|$ implements a mean aggregation, ensuring that nodes with many connections do not dominate the aggregation.

To stabilise training and facilitate gradient flow through deep architectures, each intermediate layer $\ell < L$ incorporates several architectural mechanisms. More specifically, residual connections \cite{He2016DeepRecognition} are added to mitigate the vanishing gradient problem through additive information flow across layers. In addition, layer normalisation \cite{Ba2016LayerNormalization} is employed to stabilise the distribution of representations as they evolve over the network during training. Furthermore, dropout regularisation prevents overfitting by randomly deactivating neurons during training. In contrast, the final layer $L$ differs from intermediate layers, applying only layer normalisation without an activation function or residual connection. 

This produces the final context-aware entity embedding $\mathbf{h}_v^{a,(L)} \in \mathbb{R}^{d_{\mathrm{emb}}}$, where $d_{\mathrm{emb}}$ is the embedding dimension. These learnt embeddings encode both intrinsic entity properties, derived from the initial feature encoding, and relational context, acquired through iterative message passing across the graph structure. Consequently, each embedding $\mathbf{h}_v^{a,(L)}$ captures the strategic position of entity $v$ within the market ecosystem, reflecting competitive pressures from rival services, portfolio synergies with services under the same agent's control, and network effects arising from connectivity relations.

\subsubsection{Attention-based agent-level state aggregation} \label{sect:aggregation}

Each agent GNN yields a set of context-aware embeddings $\{\mathbf{h}_v^{a,(L)}\}_{v \in V}$ for all entities in the environment. However, for decision-making, each agent requires a single, fixed-size state vector that aggregates information from the set of entities relevant to its decision-making. Since the number of entities can vary, a simple concatenation or flattening of embeddings is not feasible. While fixed pooling operations, such as mean or max aggregation, can produce a fixed-size vector, they treat all entities as equally important, regardless of their strategic relevance.

To overcome this, an attention mechanism is introduced that learns to dynamically weigh the importance of each entity in the environment. This allows the agent to focus on the most strategically relevant entities when constructing its state representation. For instance, the attention weights can learn to prioritise services in highly competitive markets during price wars, while directing focus towards capacity-constrained services when demand increases, or emphasising services that form multi-operator journeys when coordination opportunities arise. For each agent $a$, attention weights are computed by applying a learnt scoring function to each entity embedding, followed by softmax normalisation:

\begin{equation} \label{eq:attention}
    \alpha_v^a = \frac{\exp\big(\mathrm{score}(\mathbf{h}_v^{a,(L)})\big)}{\sum_{u \in V} \exp\big(\mathrm{score}(\mathbf{h}_u^{a,(L)})\big)}, \quad v \in V,
\end{equation}

where $\mathrm{score}(\cdot)$ is a learnable function that maps entity embeddings to scalar importance scores. In particular, a two-layer MLP with a ReLU activation function is used for the $\mathrm{score}$ function. The agent-level observation $\tilde{\mathbf{z}}^a$ is then computed as a weighted sum:

\begin{equation} \label{eq:agent_state}
    \tilde{\mathbf{z}}^a = \sum_{v \in V} \alpha_v^a \mathbf{h}_v^{a,(L)},
\end{equation}

yielding $\tilde{\mathbf{z}}^a \in \mathbb{R}^{d_{\mathrm{emb}}}$, a permutation-invariant representation \cite{Zaheer2017DeepSets} that dynamically weights entities according to learnt strategic relevance. Moreover, the learnt attention weights are useful diagnostics, offering insight into which entities the agent considers most relevant to make a decision at each time step.

\subsection{Relational actor-critic with heterogeneous graph embeddings} \label{sec:marl_framework}

The attention mechanism produces a context-aware state vector $\tilde{\mathbf{z}}^a$ for each agent. Next, to enable strategic decision-making with these representations in mixed competitive-cooperative multi-agent settings under partial observability and information asymmetry, the framework employs the CTDE paradigm \cite{Foerster2018CounterfactualGradients,Lowe2017Multi-agentEnvironments}. This framework gives agents access to global information during training while maintaining decentralised, privacy-preserving policies at execution time, making it well-suited for competitive market environments where operators must act independently based on local observations. Based on the MATD3 algorithm \cite{ackermann2019reducing}, the proposed method extends the standard actor-critic architecture with graph-based representation learning to capture relational environmental structure. Each agent $a$ maintains two distinct components, a decentralised actor $\pi^a_\theta$ that maps local aggregated observations $\tilde{\mathbf{z}}^a$ to actions, and two centralised critics $Q^a_{\phi_1}$, $Q^a_{\phi_2}$ that estimate action-value functions using global state information, with the minimum taken to mitigate overestimation bias.

Additionally, each agent $a$ maintains its own independent GNN ($\text{GNN}_{\psi^a}$) that processes the entity graph $\mathcal{G}_t$, with agent-specific masked features, to produce entity embeddings $\{\mathbf{h}_v^{a,(L)}\}_{v \in V}$. This design reflects the information asymmetry inherent in competitive markets where each agent observes only public information about competitors' entities while maintaining complete visibility of its own operations. In particular, the actor network $\pi^a_\theta$ processes the agent's aggregated observation $\tilde{\mathbf{z}}^a$ derived from attention-weighted pooling of entity embeddings to produce a continuous action $u^a$. The actor maximises expected returns through policy gradients, whereas the critic minimises temporal difference error. Therefore, using both to update the GNN representations can lead to conflicting gradient signals. To address this, when computing actions for environment interaction and actor updates, the aggregated observation $\tilde{\mathbf{z}}^a$, along with the representation module, is detached from the computational graph. Thus, the GNN is trained exclusively through the critic loss, which provides richer signals from the global value function, decoupling policy learning from representation learning.

The centralised critics $Q^a_{\phi_1}(\tilde{\mathbf{Z}}, \mathbf{u}),Q^a_{\phi_2}(\tilde{\mathbf{Z}}, \mathbf{u})$ take as input the joint aggregated observations $\tilde{\mathbf{Z}} = (\tilde{\mathbf{z}}^1, \ldots, \tilde{\mathbf{z}}^n)$ and joint actions $\mathbf{u} = (u^1, \ldots, u^n)$ of all agents. The critics operate on the same abstracted state space as the actors, taking as input the joint aggregated observation $\tilde{\mathbf{Z}}$ to learn a global value function conditioned on the agents' summarised local views. During critic updates, the GNN parameters $\psi^a$ are optimised only with respect to the current agent, although the critics receive the joint embedding $\tilde{\mathbf{Z}}_j$, which also contains the embeddings produced by the GNNs of other agents. The GNN parameters $\psi^a$ are jointly optimised with the critics parameters $\phi_1^a,\phi_2^a$ to minimise the temporal difference error as the expected value over samples drawn from the replay buffer $\mathcal{D}$:

\begin{equation} \label{eq:loss_critic_gnn}
    \mathcal{L}(\phi_1^a, \phi_2^a, \psi^a) = \mathbb{E}_{j \sim \mathcal{D}} \left[\sum_{i=1}^{2} \left( y^a_j - Q^a_{\phi_i}(\tilde{\mathbf{Z}}_j, \mathbf{u}_j) \right)^2\right],
\end{equation}

where $y^a_j$ is the target value computed using target networks as follows:

\begin{equation} \label{eq:target}
    y^a_j = r^a_j + (1 - d_j) \ \gamma \ \underset{i=1,2}{\text{min}} \ Q^a_{\phi'_i}(\tilde{\mathbf{Z}}_{j+1}, \pi^1_{\theta'}(\tilde{\mathbf{z}}^1_{j+1}) + \epsilon, \ldots, \pi^n_{\theta'}(\tilde{\mathbf{z}}^n_{j+1}) + \epsilon),
\end{equation}

and $\epsilon \sim \text{clip}(\mathcal{N}(0, \sigma), -c, c)$ introduces a small amount of random noise to the target policy as a regularisation term, clipped to keep the target close to the original action. The term $d_j$ indicates a terminal transition, for transitions where the episode terminates, the target $y^a_j$ is equal to the immediate reward $r^a_j$. The next-state embeddings are produced by the online representation module, as only target networks are used for the actors and critics, which already provide a stable objective in the learning process. Then, the actor is updated using the deterministic policy gradient:

\begin{equation} \label{eq:update_actor}
    \nabla_{\theta^a}J(\pi^a) = \mathbb{E}_{j\sim \mathcal{D}} \left[\nabla_{\theta^a} \pi^a(\tilde{\mathbf{z}}_j^a) \nabla_{u^a} Q_{\phi_{1}^{a}}(\tilde{\mathbf{Z}}_j, \mathbf{u}_j)|_{u^a=\pi^a(\tilde{\mathbf{z}}_j^a)}\right]
\end{equation}

with gradient detachment on the aggregated observations to preserve the separation between policy and representation learning. In addition, the actor target network is updated with a reduced frequency $d$ compared to the critic, which has been shown to reduce overestimation bias and improve sample efficiency in continuous control tasks \cite{Fujimoto2018AddressingMethods}. Finally, to further stabilise learning, target networks are updated using soft updates with interpolation parameter $\tau$. The complete training procedure of the framework called RACHE is detailed in Algorithm \ref{alg:training_algorithm}.

\begin{algorithm}[!ht]
\small
\caption{Relational Actor-Critic with Heterogeneous Graph Embeddings (RACHE)}
\label{alg:training_algorithm}
\begin{algorithmic}[1]
    \State Initialise networks $\{\pi_{\theta^a}, Q_{\phi_1^a}, Q_{\phi_2^a}, \text{GNN}_{\psi^a}\}_{a=1}^n$ with random parameters
    \State Initialise target networks $\theta{'}^{a} \leftarrow \theta^{a}, \phi{'}_1^a \leftarrow \phi_1^a, \phi{'}_2^a \leftarrow \phi_2^a$
    \State Initialise replay buffer $\mathcal{D}$
    \For{episode = $1, \ldots, M$}
        \For{$t = 1, \ldots, T$}
            \For{each agent $a = 1, \ldots, n$}
                \State Compute service embeddings: $\{\mathbf{h}_v^{a,(L)}\}_{v \in V} = \text{GNN}_{\psi^a}(\mathcal{G}_t)$ \Comment{With masked features}
                \State Compute aggregated observation $\tilde{\mathbf{z}}^a_t$ via attention using Eq.~\ref{eq:attention}-\ref{eq:agent_state}
                \State Select action: $\mathbf{u}_t^a = \pi_{\theta^a}(\tilde{\mathbf{z}}_t^a) + \epsilon$, $\epsilon \sim \mathcal{N}(0,\sigma)$ \Comment{Stop gradient}
            \EndFor
            \State Execute joint action $\mathbf{u}_t$, observe rewards $\{r_t^a\}_{a=1}^n$ and next observation $\mathbf{Z}_{t+1}$
            \State Store transition $(\mathbf{Z}_t, \mathbf{u}_t, \{r_t^a\}_{a=1}^n, \mathbf{Z}_{t+1}, d_t)$ in $\mathcal{D}$ \Comment{Raw entity features}
            \If{$|\mathcal{D}| \geq $ batch size}
                \For{each agent $a = 1, \ldots, n$}
                    \State Sample minibatch $(\mathbf{Z}_j, \mathbf{u}_j, \{r_j^a\}_{a=1}^n, \mathbf{Z}_{j+1}, d_t)$ from $\mathcal{D}$
                    \State Compute aggregated observation $\tilde{\mathbf{z}}_{j+1}^{a}$ for all agents \Comment{See Lines 7-8}
                    \State Compute target as shown in Eq.~\ref{eq:target}
                    \State Update critic and GNN using Eq.~\ref{eq:loss_critic_gnn}
                    \State Update actor using the sampled policy gradient as in Eq.~\ref{eq:update_actor}
                    \Statex \Comment{Stop gradient}
                \EndFor
                \State Update critic target network: $\phi{'}_i^{a} \leftarrow \tau\phi_i^a + (1 - \tau) \phi{'}_i^{a}, \ i \in \{1, 2\}$
                \If{$t \bmod d = 0$}
                    \State Update actor target network: $\theta{'}^{a} \leftarrow \tau\theta^a + (1 - \tau) \theta{'}^{a}$
                \EndIf
            \EndIf
        \EndFor
    \EndFor
\end{algorithmic}
\end{algorithm}

\section{Experimental settings and results} \label{sec:experiments}

This section evaluates the effectiveness of the proposal in learning strategic pricing policies in two different scenarios. First, Section \ref{sec:environment} describes the simulation environment, and Section \ref{sec:scenarios} the experimental scenarios. Subsequently, Section \ref{sec:algorithms} presents the algorithms selected for evaluation. Then, Section \ref{sec:implementation_details} details the implementation settings. Later, Section \ref{sec:performance_analysis} analyses the comparative performance of the evaluated algorithms and finally, Section \ref{sec:further_studies} examines key architectural components of RACHE.

\subsection{Simulation environment} \label{sec:environment}

All experiments are conducted using the simulator RailPricing-RL \cite{Villarrubia-Martin2025DynamicLearning}, a parameterisable reinforcement learning environment designed to model high-speed railway networks featuring dynamic pricing, multi-operator itineraries, and diverse passenger behaviour. The environment employs microscopic modelling of passenger decision-making using RUM, where passengers can choose not to travel if no itinerary meets their utility threshold, resulting in a non-zero sum game. It supports MARL algorithms through compatibility with standard RL interfaces. In this environment, each service is scheduled for a specific date. 
However, the simulation operates over a booking horizon that begins when the first passenger attempts to buy a ticket. This horizon is divided into discrete time steps, each representing a single day. During these steps, agents set prices, and passengers, each arriving on a specific day sampled from their segment's booking distribution, make purchasing decisions. Now, the observation space, action space, and reward function are described.

\paragraph{\textbf{Observation space}}
Each agent observes all services in the network. For each service the observation includes static attributes which are the train service provider, the corridor (a set of stations within a region), the line (a sequence of stations in the corridor), the time slot, and the rolling stock, encoded as indices, as well as dynamic pricing data containing the prices per service-seat combination, the ticket sales and the cumulative service revenue. Nevertheless, the ticket sales and the cumulative service revenue are private, so agent $a$ observes them only for its own services, while the corresponding features of competitor services are masked with zeros to preserve partial observability.

\paragraph{\textbf{Action space}}
The action space defines the set of decisions available to agents for modifying ticket prices in their respective services. Each agent selects a continuous action $\alpha_{v,t}^c \in [-1, 1]$ per service-seat combination, representing a normalised percentage price adjustment applied to the current price:

\begin{equation*}
    p_{v,t+1}^c = p_{v,t}^c \cdot \left(1 + \alpha_{v,t}^c \cdot \frac{\beta}{100}\right),
\end{equation*}

where $\beta = 25$ is the default scaling factor and prices are clipped to $[0, \infty)$ to prevent negative values.

\paragraph{\textbf{Reward}}
At each time step $t$, agent $a$ receives a reward equal to the incremental daily revenue across its operated services:

\begin{equation*}
    r_t^a = \sum_{v \in V^a} \rho_t^v - \rho_{t-1}^v,
\end{equation*}

where $\rho_t^v$ denotes the cumulative revenue of service $v$ up to time step $t$ and $V^a \subseteq V$ is the set of services operated by agent $a$. This formulation assumes no associated costs for providing services, so the reward captures revenue changes only. Thus, the objective is to maximise its expected total revenue over the course of an episode.

\subsection{Scenarios} \label{sec:scenarios}

\begin{figure*}[!ht]
    \begin{center}
        \begin{subfigure}{0.49\textwidth}
            \includegraphics[width=1\linewidth]{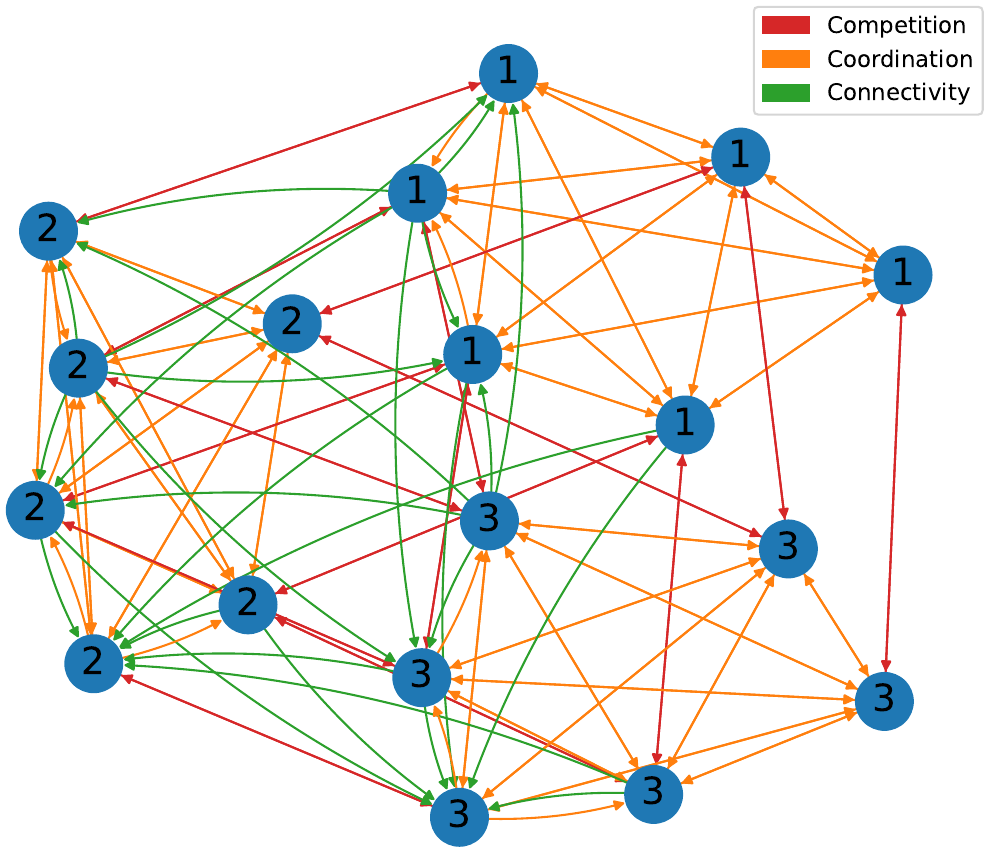}
            \caption{Base.}
            \label{fig:supply_base}
        \end{subfigure}
        \begin{subfigure}{0.49\textwidth}
            \includegraphics[width=1\linewidth]{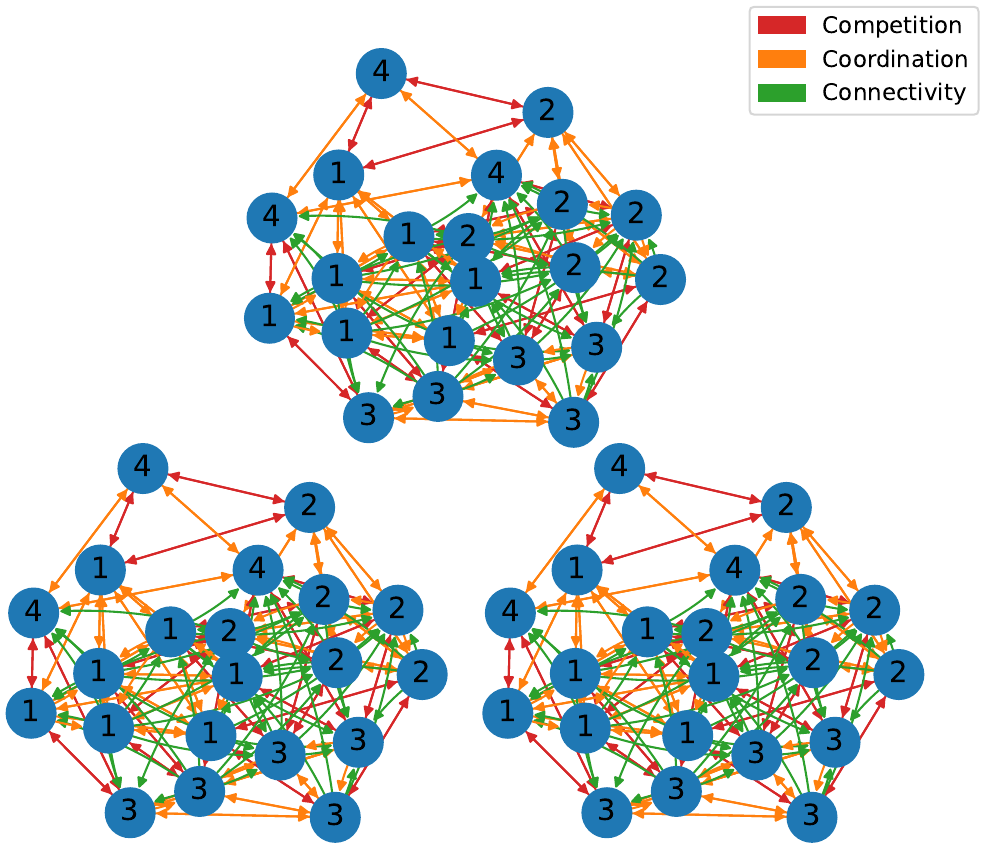}
            \caption{Large.}
            \label{fig:supply_large}
        \end{subfigure}
    \end{center}
    \caption{Entity graph representations for the experimental scenarios. Each node corresponds to a train service labelled by its operating agent, and edges encode competition, coordination, and connectivity relations.}
    \label{fig:service_graph_scenarios}
\end{figure*}

The experiments evaluate the performance of the algorithms in two scenarios of increasing market complexity. These scenarios employ distinct supply configurations and a common demand pattern to assess algorithm behaviour under varying levels of market structure. The supply configuration of the first scenario features 18 services scheduled on the same date, operated by three agents, each one with six services, illustrated in Figure~\ref{fig:supply_base}. The second scenario expands it to 63 services, consisting of a daily schedule of 21 services repeated on three consecutive days. In addition, it introduces an additional agent as shown in Figure~\ref{fig:supply_large}. In both scenarios, agents compete directly in overlapping markets and can cooperate through multi-operator connecting itineraries. Agents are heterogeneous, controlling different sets of services with distinct action spaces. The specific characteristics of each scenario are described below.

\paragraph{\textbf{Base}}
This scenario introduces a varied passenger market across five stations and seven origin-destination markets, operated by three agents, each controlling six services on a single day with a total of 18 services. Four passenger segments named business, commuter, student and leisure coexist with distinct utility functions and temporal preferences. On the one hand, business travellers are a user group characterised by inelastic demand, with strong preferences for convenient morning arrival and departure times, and commuters exhibit high penalties for travel time and transfers. On the other hand, students show high price sensitivity with flexible timing preferences and book tickets close to the departure, whereas leisure passengers tend to book earlier and over a wider window with greater diversity in the arrival time. Episodes last seven days and generate an average of 980 potential passengers across all markets. Moreover, all services offer two seat types, basic and premium, with a capacity of 32 and eight seats per service respectively, forming a joint action space of 36 dimensions. Finite seat availability introduces capacity management constraints requiring agents to adapt their pricing strategies in response to competition and inventory levels.

\paragraph{\textbf{Large}}
This scenario expands market complexity by introducing a fourth competing agent, resulting in a total of 63 services, 21 per day over three consecutive days, across six origin-destination markets with an increased number of passengers. The four agents are asymmetric in portfolio size, with the largest controlling a total of 21 services, and the smallest operating only nine. The same four passenger segments coexist with the same utility functions and temporal preferences described above. Episodes span ten days and generate an average of approximately 1,420 potential passengers across all markets. As in the first scenario, all services offer basic and premium seat types, with a capacity of 13 and four seats per service respectively, and a joint action space of 126 dimensions, to account for capacity management constraints.

Finally, Table \ref{tab:scenarios} reports the statistics of the entity graph for each scenario. In both, the coordination relation is the densest one, as it connects every service scheduled on the same date provided by the same agent. However, the other relations in the Large scenario, competition and connectivity, are more represented with the addition of a fourth agent.

\begin{table}[!ht]
  \caption{Statistics of the entity graph for each scenario. Competition and coordination edges are symmetric and counted in both directions, whereas connectivity edges are directed.}
  \small
  \begin{tabular*}{\textwidth}{@{\extracolsep\fill}lcc}
      \toprule
      \textbf{Statistic} & \textbf{Base} & \textbf{Large} \\
      \midrule
      Agents & 3 & 4 \\
      Nodes (services) & 18 & 63 \\
      Competition edges & 30 (21\%) & 162 (26\%) \\
      Coordination edges & 90 (61\%) & 294 (47\%) \\
      Connectivity edges & 27 (18\%) & 168 (27\%) \\
      Total edges & 147 & 624 \\
      \bottomrule
  \end{tabular*}
  \label{tab:scenarios}
\end{table}

\subsection{Benchmark algorithms and baselines} \label{sec:algorithms}

To evaluate the approach and its contribution of entity-based graph modelling, RACHE is compared against a representative selection of MARL algorithms covering deterministic and stochastic policies, with and without relational structure. The evaluated algorithms are described below:

\begin{itemize}
    \item \textbf{Random:} A non-learning baseline that selects pricing actions uniformly at random from the action space, establishing a lower bound on performance.
    \item \textbf{Static pricing:} A revenue management strategy in which fixed pricing is used to assess whether dynamic pricing policies are valuable.
    \item \textbf{MADDPG \cite{Lowe2017Multi-agentEnvironments}:} A deterministic actor-critic algorithm extending DDPG \cite{Lillicrap2016ContinuousLearning} to multi-agent settings via CTDE, included as a baseline for deterministic policy methods in mixed cooperative-competitive environments.
    \item \textbf{MATD3 \cite{ackermann2019reducing}:} A deterministic actor-critic that extends MADDPG with twin critics, delayed policy updates, and target policy smoothing to mitigate overestimation bias. This is the algorithm the framework is based on.
    \item \textbf{MAAC \cite{Iqbal2019Actor-attention-criticLearning}:} A stochastic multi-agent extension of SAC \cite{Haarnoja2018SoftActor} incorporating attention to dynamically weight the relevance of other agents when computing centralised Q-values to assess agent-level attention without explicit graph structure.
    \item \textbf{GA-AC \cite{Liu2020Multi-AgentNetwork}:} A graph-based stochastic actor-critic in which decision-making agents are graph nodes and interactions are modelled by the G2ANet two-stage attention mechanism, combining hard gating with soft scaled dot-product attention. It contrasts the agent-based graph paradigm with the entity-graph approach of RACHE.
\end{itemize}

These algorithms are evaluated in the scenarios described previously to assess whether explicit relational reasoning provides benefits compared to flat observation encoders, whether modelling operational entities in an entity graph is preferable to modelling decision-making agents as graph nodes, and how deterministic policy methods compare with stochastic alternatives in this setting. All baseline implementations, except for the MAAC algorithm for which the original repository has been used, are publicly available at: \url{https://github.com/Kinrre/RelationalRailPricing-RL}.

\subsection{Implementation details} \label{sec:implementation_details}

For a standardised comparison, the default hyperparameters of each algorithm were adopted without additional tuning. A complete list of the hyperparameters is provided in \ref{sec:hyperparameters}, and all experiments were conducted on an Intel i7-13700KF CPU and an NVIDIA GeForce RTX 4080 16GB GPU. To encourage initial exploration and diversify the state-action space, each algorithm was first trained with a random policy for 1,000 episodes. The models were trained for one million steps with a replay buffer of the same size. Furthermore, policy updates were performed every two Q-network updates in the case of the MATD3 and RACHE algorithms. Finally, it is worth noting that the RACHE framework employs a two-layer R-GCN for relational context propagation across the entity graph.

In the experiments, to facilitate stable training, normalisation was applied to continuous features of the observation space, such as prices, ticket sales and cumulative service revenue. In particular, it was performed after masking the private data of each agent, which can still be identified through the operator feature. The normalised observation feature $\hat{\mathbf{z}}_t$ at time step $t$ was computed as follows:

\begin{equation*}
    \hat{\mathbf{z}}_t = \frac{\mathbf{z}_t - \mu_t}{\sqrt{\sigma_t^2 + \epsilon}},
    \label{eq:normalization}
\end{equation*}

where $\mathbf{z}_t$ is the raw observation at time $t$, $\mu_t$ is the running mean of each individual continuous feature up to time $t$, $\sigma_t^2$ is the running variance of each individual feature up to time $t$, and $\epsilon$ is a small constant used for numerical stability. For the reward signal, representing the daily revenue, reward scaling was employed instead of standard normalisation. Rewards were divided by a constant factor of 1,000 in both scenarios to account for high revenue magnitudes. This approach preserves the direct relationship between revenue and the reward signal, maintaining positive signs and relative magnitudes while preventing numerical instability from large gradient updates. As a result, the value function receives a more consistent learning signal with a clearer distinction between more profitable and less profitable actions.

\subsection{Performance comparisons and analysis} \label{sec:performance_analysis}

\begin{figure}[!ht]
    \centering
    \includegraphics[width=\textwidth]{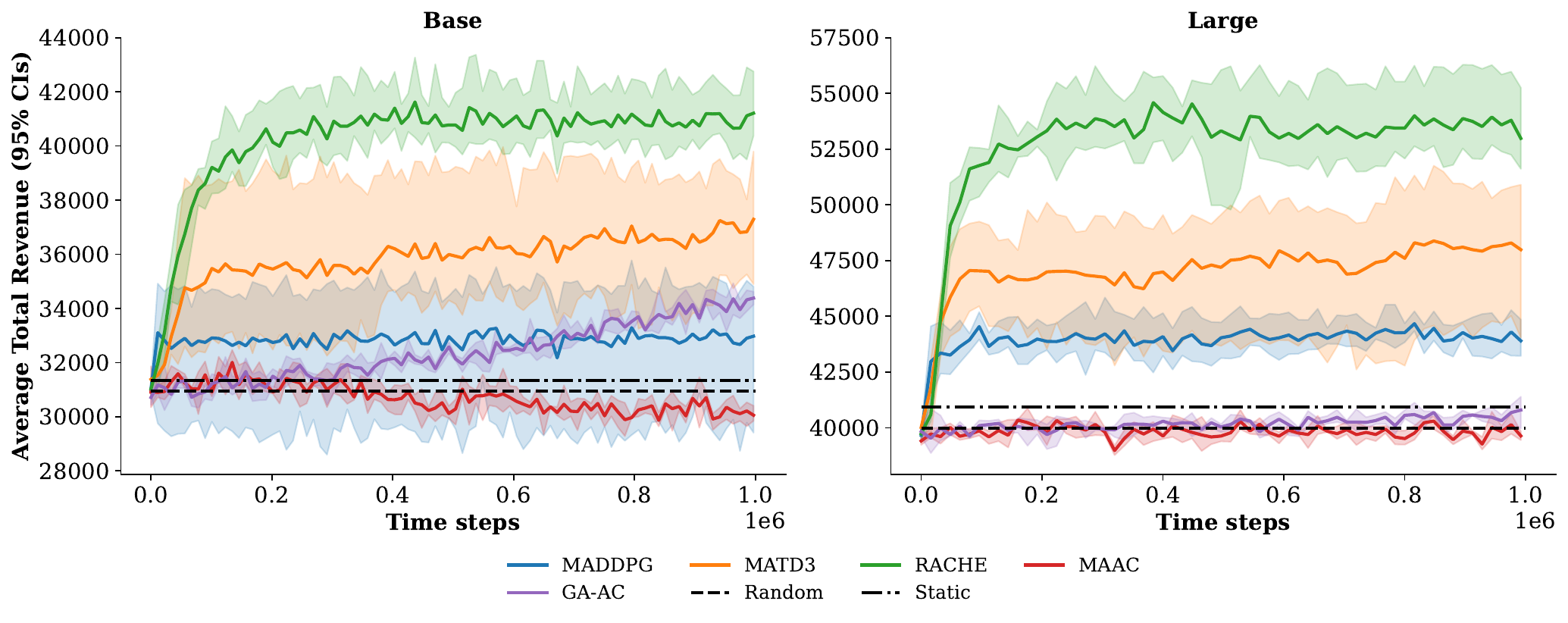}
    \caption{Average total revenue obtained at training for the two scenarios.}
    \label{fig:training_curves}
\end{figure}

This section analyses the performance of the evaluated algorithms in both scenarios. Table \ref{tab:total_revenue} summarises the total revenue obtained during the evaluation at the end of training using the last checkpoint of the models, and Figure \ref{fig:training_curves} illustrates the training dynamics across all scenarios.

\begin{table}[!ht]
    \caption{Total revenue obtained in the evaluation using the last checkpoint for the different scenarios. The mean and the stratified bootstrap confidence intervals at 95\% were computed over three independent runs.}
    \small
    \begin{tabular*}{\textwidth}{@{\extracolsep\fill}lcc}
        \toprule
        \textbf{Algorithm} & \textbf{Base} & \textbf{Large} \\
        \midrule
        Random & $30961 \ [30832, 31092]$ & $40031 \ [39937, 40115]$ \\
        Static pricing & $31309 \ [31284, 31335]$ & $40950 \ [40908, 40992]$ \\
        MADDPG & $32688 \ [31641, 33675]$ & $43902 \ [43427, 44402]$ \\
        MATD3 & $37028 \ [36312, 37739]$ & $48206 \ [46930, 49425]$ \\
        MAAC & $31254 \ [30445, 32093]$ & $41220 \ [40755, 41675]$ \\
        GA-AC & $34446 \ [34285, 34596]$ & $40479 \ [40388, 40566]$ \\
        RACHE (Ours) & $\mathbf{41036 \ [40596, 41509]}$ & $\mathbf{53831 \ [53113, 54604]}$ \\
        \bottomrule
    \end{tabular*}
    The highest values are indicated in bold.
    \label{tab:total_revenue}
\end{table}

First, with respect to the non-learning baselines, static pricing, that keeps a fixed pricing by not modifying the initial price point for each service specified by the scenarios, performs better than the random baseline, indicating that modifying prices randomly from the action space is a worse strategy. Among the learning-based algorithms, the RACHE framework achieves the highest revenue in both scenarios, followed by MATD3, which presents a substantially higher variance. This supports that the relational state representation learnt over the entity graph provides a meaningful inductive bias. As expected, regarding the deterministic baselines, MATD3 surpasses MADDPG, since the combination of twin critics, delayed policy updates, and target policy smoothing helps to achieve a higher revenue.

In the case of the stochastic alternatives, MAAC fails to learn an effective pricing policy, performing comparably to or even worse than the random baseline during training in the first scenario. Finally, regarding GA-AC, the agent-based graph baseline learns a meaningful policy that surpasses MADDPG in the Base scenario, with a very low variance. However, in the large one, its performance fails to improve over the course of training. This is due to the fact that, by abstracting each operator as a single node, GA-AC cannot capture the fine-grained service-level relations. The entity graph addresses precisely this limitation by treating services rather than operators as the primary vertices, whose granularity scales with the market structure.

\begin{figure}[!ht]
    \centering
    \includegraphics[width=\textwidth]{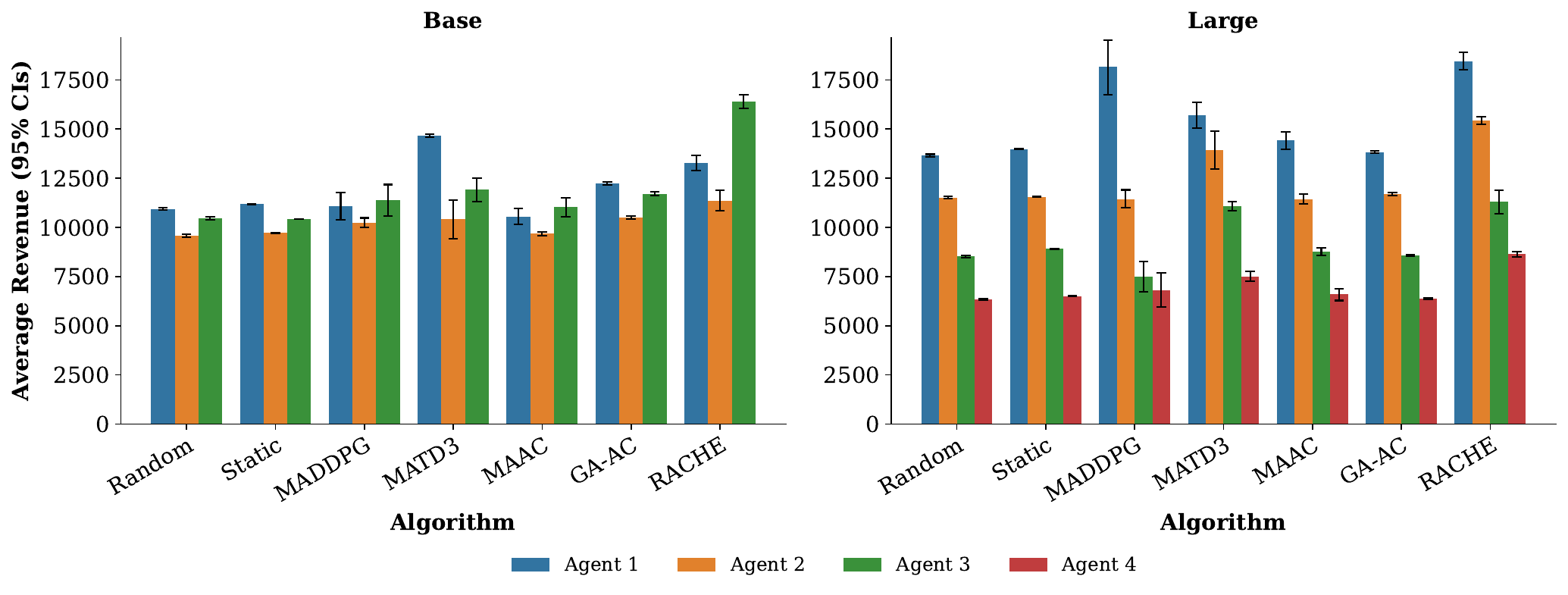}
    \caption{Average revenue obtained at evaluation using the last checkpoint for each agent.}
    \label{fig:revenue_per_agent}
\end{figure}

To account for which agent the total revenue comes from, Figure \ref{fig:revenue_per_agent} shows the distribution of the total revenue of each individual agent in both scenarios. Foremost, it can be observed in the Base scenario that the total revenue is more evenly distributed among the agents, whereas in the Large scenario, it is imbalanced because of the asymmetric portfolio sizes. In both cases, the RACHE framework increases the revenue of every agent with respect to almost all baselines, except, for instance, the first agent of the MATD3 algorithm in the former. More precisely, the increase of total revenue in the first scenario can be attributed mainly to the third agent, although, in the second, the improvement is more uniformly distributed across all agents.

\begin{figure}[!ht]
    \centering
    \includegraphics[width=\textwidth]{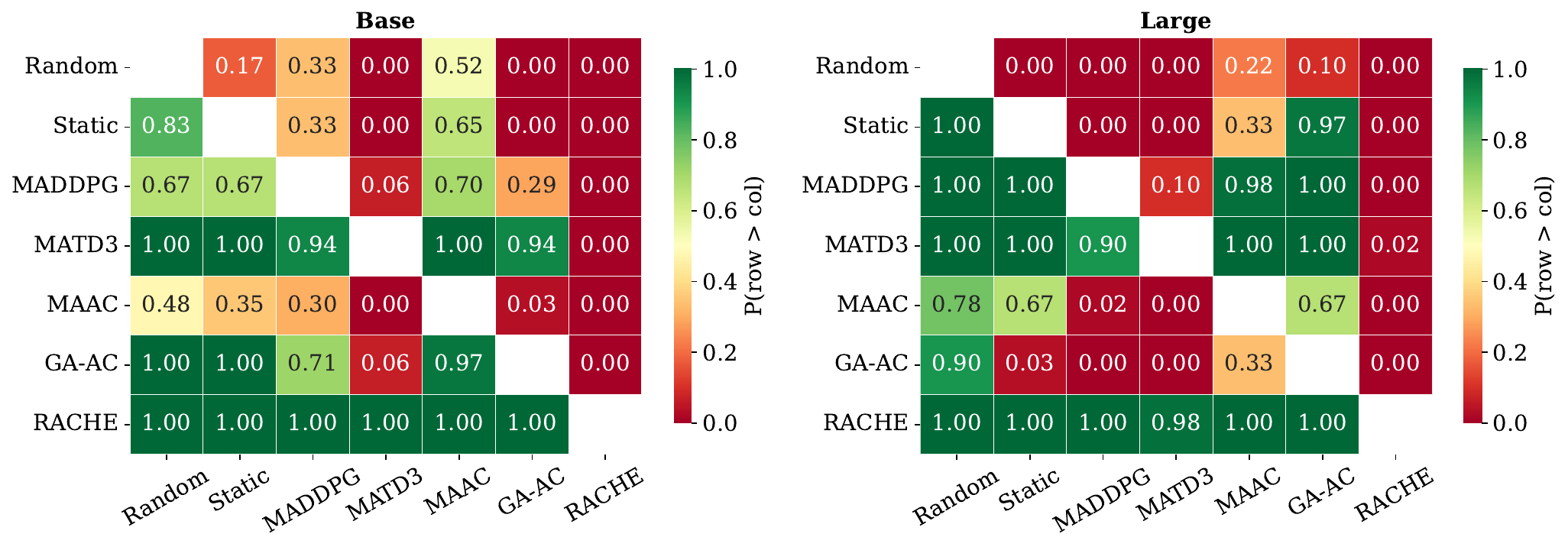}
    \caption{Probability of improvement matrix using the Mann--Whitney $U$ statistic.}
    \label{fig:prob_improvement}
\end{figure}

In addition, as suggested by \cite{Agarwal2021}, Figure \ref{fig:prob_improvement} presents a pairwise matrix with the probability of improvement, that is, the probability that algorithm $X$ obtains a higher total reward than algorithm $Y$. This is calculated from the Mann--Whitney $U$ statistic collected across the three seeds for a scenario $m$ given $N$ and $K$ evaluation episodes respectively, defined as:

\begin{equation*}
    P(X_m > Y_m) = \frac{1}{NK} \sum_{i=1}^{N} \sum_{j=1}^{K} S(x_{m,i}, y_{m,j}) \quad \text{where} \quad S(x,y) = \begin{cases} 
        1, & \text{if } y < x, \\ 
        \frac{1}{2}, & \text{if } y = x, \\ 
        0, & \text{if } y > x. 
    \end{cases}
\end{equation*}

The results indicate that the RACHE framework obtains a higher total reward than the rest of the considered baselines.

\begin{figure}[!ht]
    \centering
    \includegraphics[width=\textwidth]{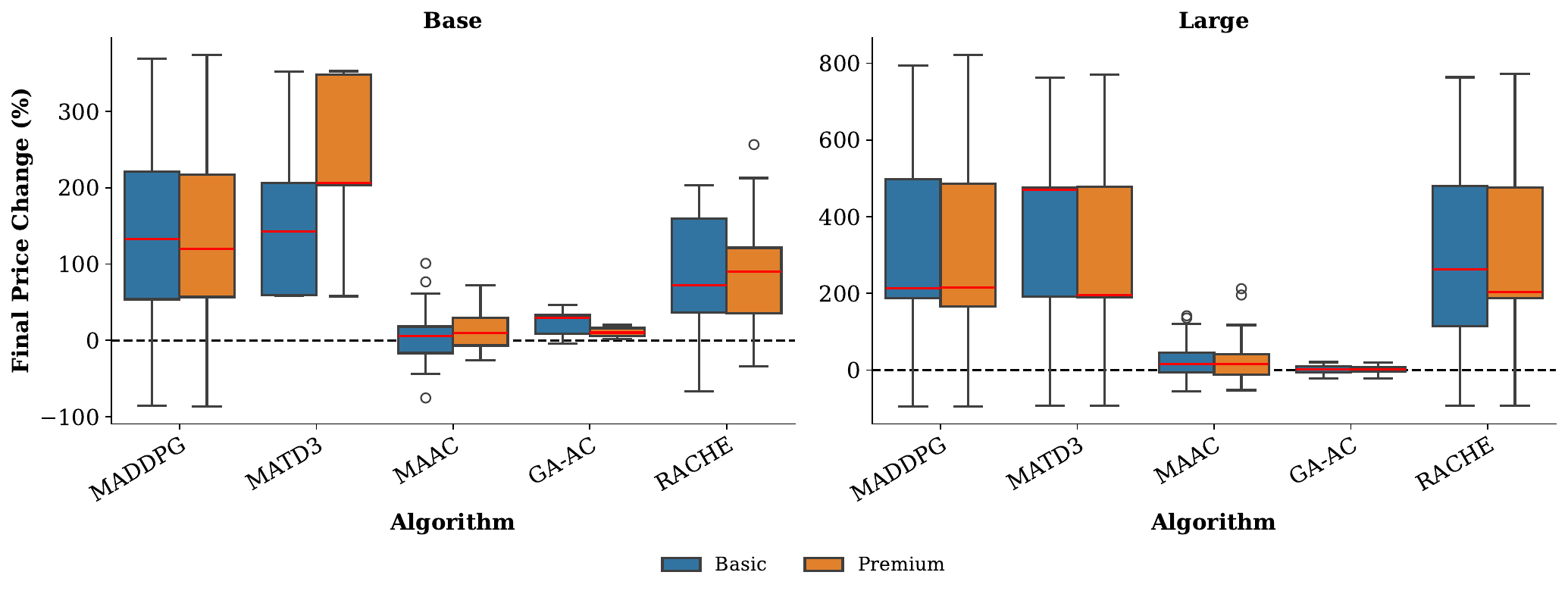}
    \caption{Distribution of the final price change per seat type relative to the initial fare.}
    \label{fig:price_changes}
\end{figure}

Next, the effect of the learnt policies on the final prices at the end of the episode, relative to the initial fares, is analysed together with its impact on the percentage of passengers travelling. Figure \ref{fig:price_changes} shows the distribution of the final price change per seat type, and Table \ref{tab:passengers_travelling} details the corresponding percentages. As shown, the deterministic algorithms increase prices more than the stochastic ones. In fact, in the Base scenario, GA-AC obtains a higher revenue than MADDPG with smaller price increments, as well as RACHE compared to MATD3 in both scenarios, indicating that larger price increases do not necessarily translate into higher revenue. In contrast, the deterministic algorithms reduce the percentage of passengers travelling more than the stochastic ones, as a consequence of their larger price increases, but still RACHE obtains a higher revenue while retaining a larger number of passengers travelling compared to MATD3.

\begin{table}[!t]
    \caption{Percentage of passengers travelling in the evaluation using the last checkpoint for the different scenarios. The mean and the stratified bootstrap confidence intervals at 95\% were computed over three independent runs.}
    \small
    \begin{tabular*}{\textwidth}{@{\extracolsep\fill}lcc}
        \toprule
        \textbf{Algorithm} & \textbf{Base} & \textbf{Large} \\
        \midrule
        Random & $70.15 \ [69.96, 70.34]$ & $63.07 \ [62.99, 63.16]$ \\
        Static pricing & $\mathbf{70.67 \ [70.50, 70.84]}$ & $\mathbf{63.22 \ [63.12, 63.31]}$ \\
        MADDPG & $63.16 \ [62.91, 63.42]$ & $52.77 \ [52.27, 53.29]$ \\
        MATD3 & $61.09 \ [59.94, 62.18]$ & $52.08 \ [51.35, 52.78]$ \\
        MAAC & $69.53 \ [69.23, 69.82]$ & $62.15 \ [61.88, 62.45]$ \\
        GA-AC & $69.48 \ [69.28, 69.68]$ & $63.10 \ [62.99, 63.20]$ \\
        RACHE (Ours) & $63.46 \ [62.03, 64.81]$ & $53.55 \ [53.04, 54.04]$ \\
        \bottomrule
    \end{tabular*}
    The highest values are indicated in bold.
    \label{tab:passengers_travelling}
\end{table}

Lastly, Table \ref{tab:parameter_wall_time} reports the number of parameters for each model and the total training time including all the seeds of each algorithm in hours. Notably, the scalability of the proposed framework can be attributed to two points: first, the processing of a compact fixed-size embedding of each agent, which decreases the input size of the centralised critics, and second, to the message passing paradigm whose parameter weights are shared across all nodes and, therefore, independent of the number of services in the graph.

\begin{table}[!ht]
    \caption{Parameter count of each algorithm and total wall training time using three independent runs for the different scenarios.}
    \small
    \begin{tabular*}{\textwidth}{@{\extracolsep\fill}lcccc}
        \toprule
        & \multicolumn{2}{c}{\textbf{Base}} & \multicolumn{2}{c}{\textbf{Large}} \\
        \cmidrule{2-5}
        & Parameter Count & Wall time & Parameter Count & Wall time \\
        \midrule
        MADDPG & $\mathbf{1,318,695}$ & $6.2 \ \text{hours}$ & $5,851,778$ & $9.0 \ \text{hours}$ \\
        MATD3 & $2,208,810$ & $6.2 \ \text{hours}$ & $10,374,790$ & $9.1 \ \text{hours}$ \\
        MAAC & $1,759,000$ & $7.0 \ \text{hours}$ & $5,152,212$ & $10.0 \ \text{hours}$ \\
        GA-AC & $4,287,825$ & $6.3 \ \text{hours}$ & $10,094,856$ & $9.1 \ \text{hours}$ \\
        RACHE (Ours) & $1,887,729$ & $\mathbf{5.8 \ \textbf{hours}}$ & $\mathbf{2,983,466}$ & $\mathbf{7.4} \ \textbf{hours}$ \\
        \bottomrule
    \end{tabular*}
    The lowest values are indicated in bold.
    \label{tab:parameter_wall_time}
\end{table}

\subsection{Further studies} \label{sec:further_studies}

This section examines key architectural components in more depth. First, Section \ref{sec:relation_ablation} analyses the importance of each relation type in the entity graph. Subsequently, Section \ref{sec:rgcn_depth} examines how the depth of message passing affects agent performance. Then, Section \ref{sec:uniform_attention} evaluates the attention-based pooling contribution. Next, Section \ref{sec:detach_ablation} investigates the effect of decoupling representation learning from the actor in the training dynamics. Finally, Section \ref{sec:embeddings} analyses the information encoded by the learnt graph embeddings.

\subsubsection{What is the contribution of each relation type in the entity graph?} \label{sec:relation_ablation}

An evaluation of the importance of each relation was carried out, in which each relation type was removed from the entity graph in order to assess its relative contribution. To this end, three variants of the framework are compared versus the default configuration, in which all three relation types are present.

\begin{figure}[!ht]
    \centering
    \includegraphics[width=1\textwidth]{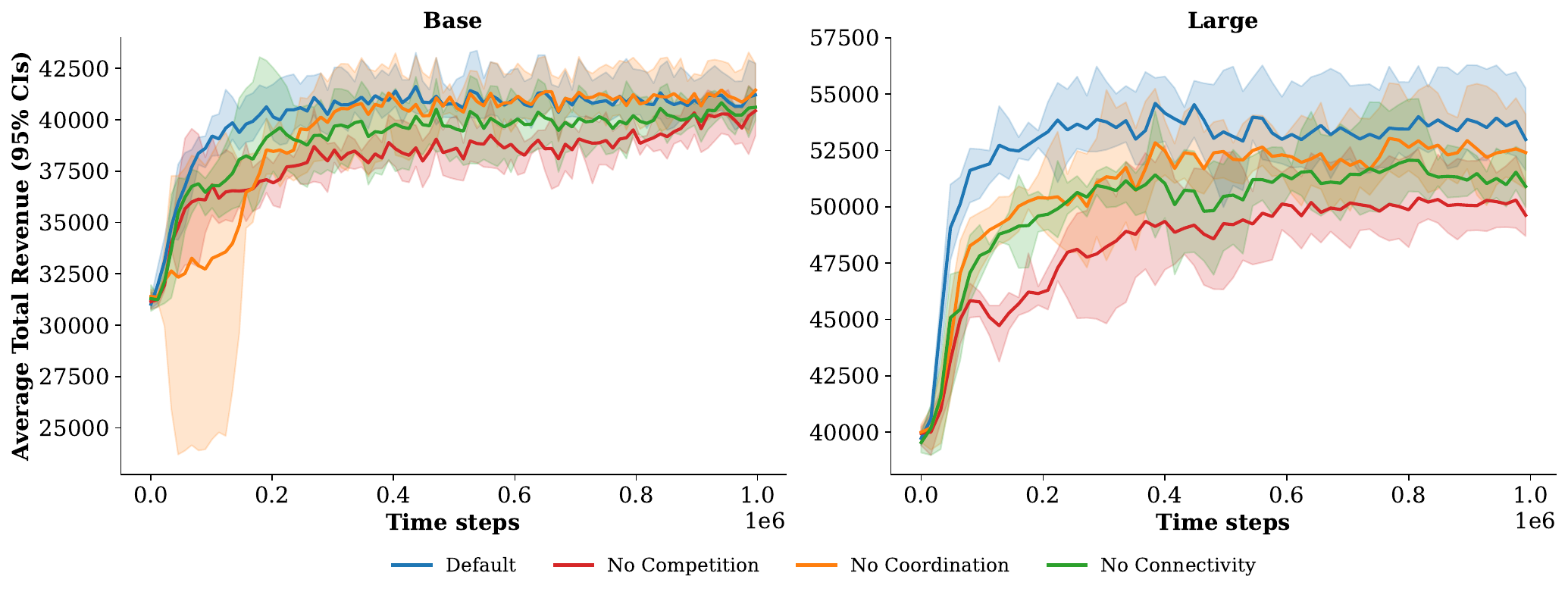}
    \caption{Ablation study comparing average total revenue with each relation type removed.}
    \label{fig:relation_ablation}
\end{figure}

Figure \ref{fig:relation_ablation} shows the average total revenue during training when each relation type is removed in both scenarios. The default configuration with all relations results in the highest total revenue, confirming that the three relations contribute to the agent's decision-making. More specifically, removing the competition relation produces the largest drop in performance, followed by the connectivity relation, which is expected, since they capture direct competition and potential cooperation between agents. In contrast, removing the coordination relation, inspired by agent-based graphs, which only encodes whether two services are operated by the same agent, leads to the smallest performance gap. This indicates that the strategic value of connecting two services derives from their market relationship rather than shared ownership.

\subsubsection{How does message passing depth affect agent performance?} \label{sec:rgcn_depth}

To evaluate the impact of network depth on agent performance, an ablation study was conducted comparing variants with one, two, and three R-GCN layers. The depth of the GNN determines how information is aggregated for each service: a single-layer GNN aggregates information only from adjacent services, whereas a two-layer or three-layer GNN expands it to two-hop or three-hop neighbourhoods, respectively.

\begin{figure}[!ht]
    \centering
    \includegraphics[width=1\textwidth]{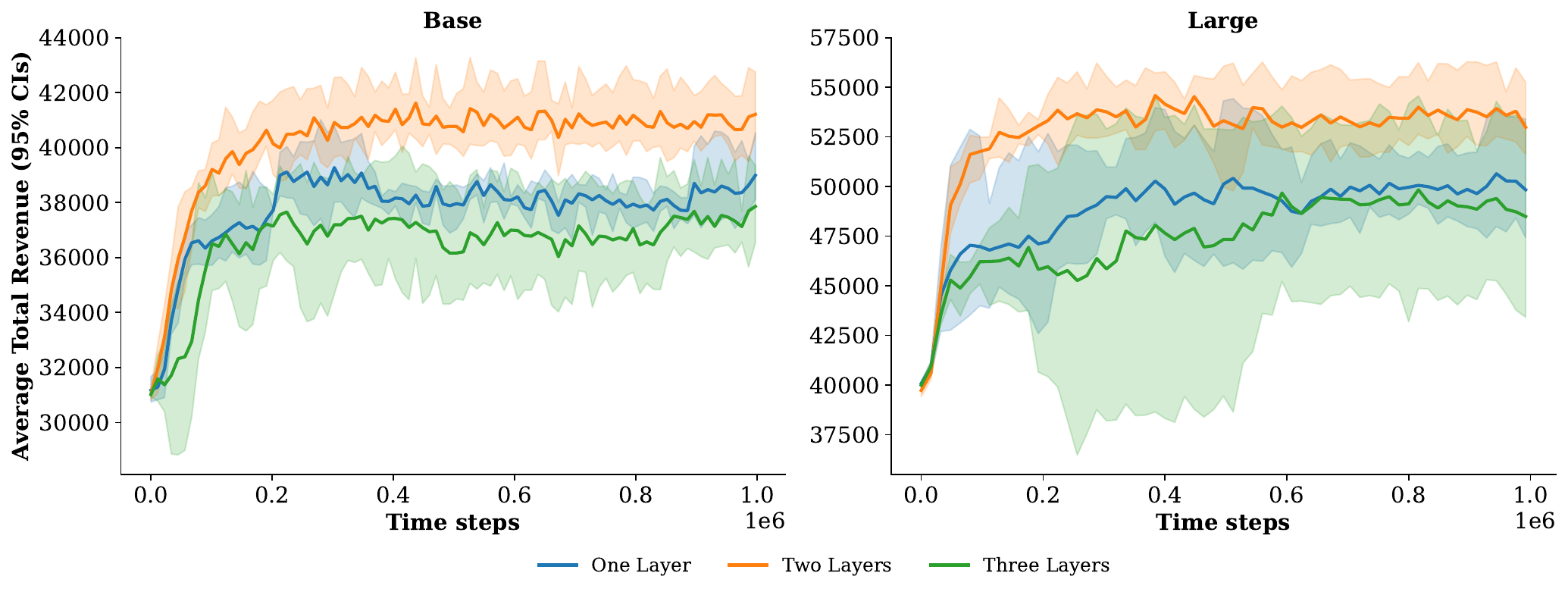}
    \caption{Ablation study comparing average total revenue for one, two, and three layers.}
    \label{fig:depth_ablation}
\end{figure}

Figure \ref{fig:depth_ablation} illustrates the average total revenue during training for each depth setting. The two-layer configuration achieves the best performance in both scenarios, followed by the one-layer and three-layer variants. This can be attributed to the fact that it captures both, direct and connecting services with one transfer, in contrast to the one-layer variant, which only captures direct services, as price-sensitive passengers are willing to accept longer journeys or transfers if the total fare is lower. In addition, the three-layer variant exhibits the worst performance, which can suggest a smoothing effect on the features of vertices, making them harder to distinguish and less meaningful for decision-making \cite{Li2018}.

\subsubsection{How does attention-based pooling compare to uniform aggregation?} \label{sec:uniform_attention}

The attention mechanism serves as a dynamic weighting function that aggregates service-level embeddings into agent-level observations. To assess the contribution of the learnt attention mechanism to agent performance, the framework with learnt attention weights was compared against a variant employing uniform attention weights. In the uniform attention variant, all services receive an equal weight during the aggregation process, resulting in the computation of a mean service embedding, rather than a weighted combination.

\begin{figure}[!ht]
    \centering
    \includegraphics[width=1\textwidth]{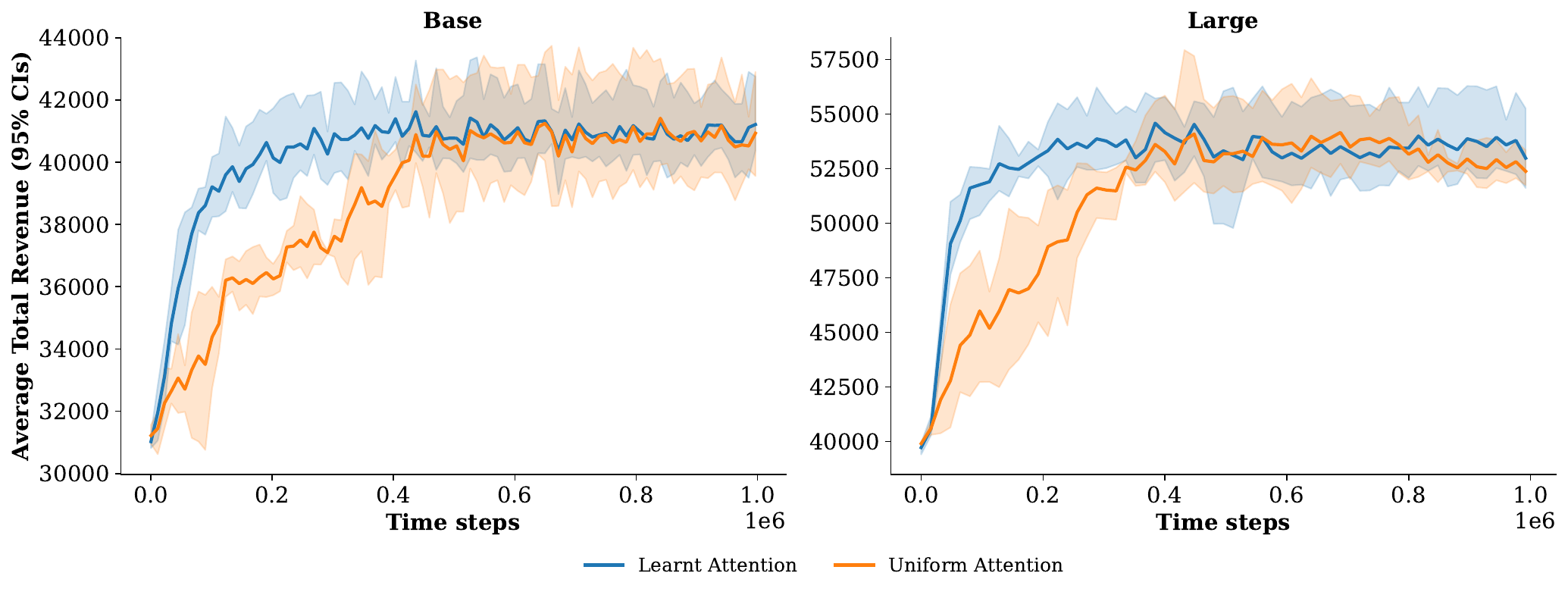}
    \caption{Ablation study comparing average total revenue of learnt attention weights against uniform attention weights.}
    \label{fig:attention_ablation}
\end{figure}

Figure \ref{fig:attention_ablation} presents the average total revenue during training for both approaches. The results show that the uniform attention variant leads to a slower learning process and, hence, to a worse sample efficiency compared to the learnt attention method. Moreover, an interesting finding is that, once enough experience has been collected, both variants converge to a similar asymptotic performance. As a result, the multi-layer R-GCN compensates for the lack of dynamic weighting by encoding later the relevant relational context directly into the entity embeddings.

\begin{figure}[!ht]
    \centering
    \includegraphics[width=1\textwidth]{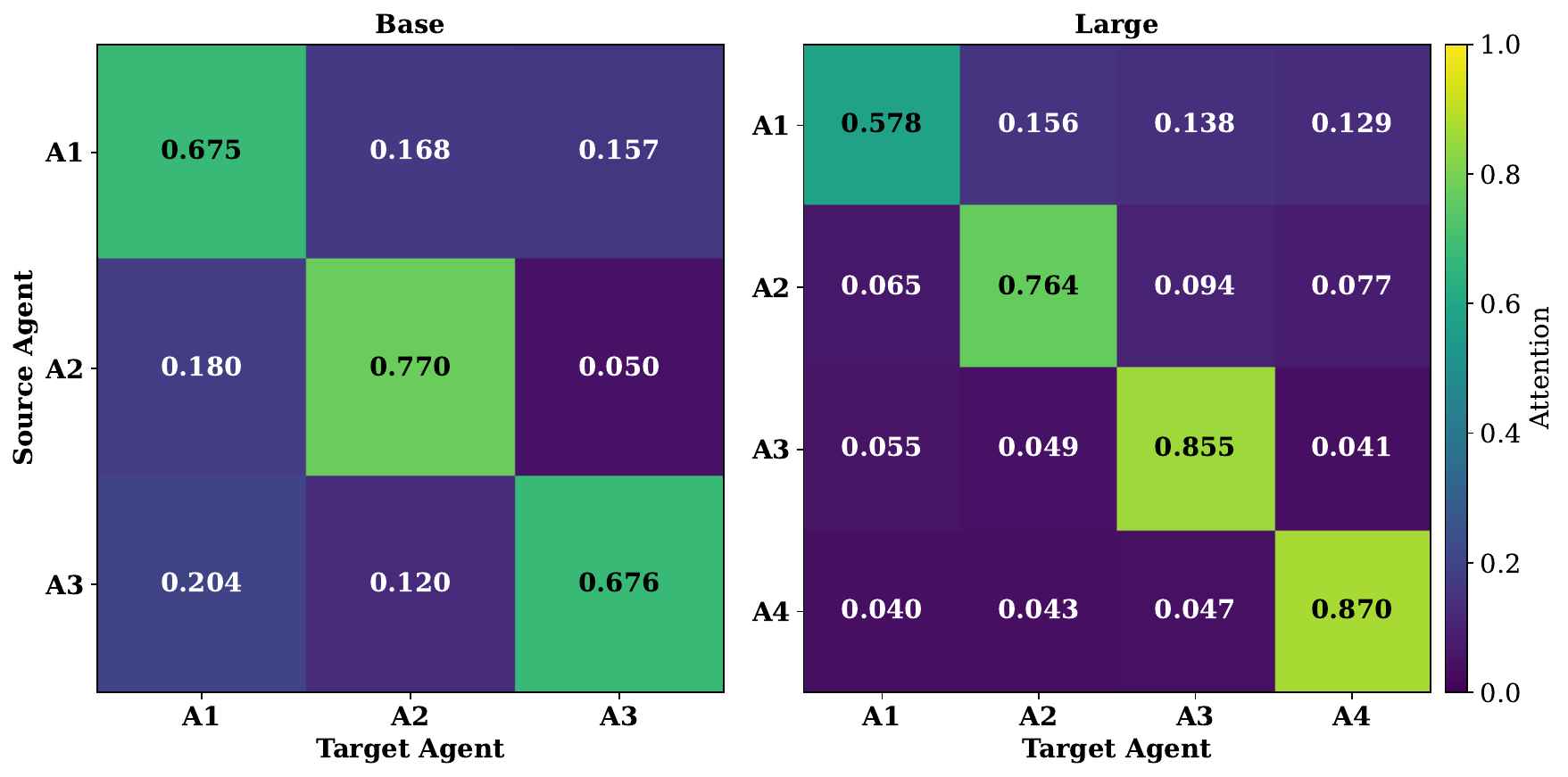}
    \caption{Normalised mean cross-attention weights between agents at training step 200K.}
    \label{fig:cross_attention}
\end{figure}

In addition, a visual representation of the cross-attention is shown in Figure \ref{fig:cross_attention}, where the coordinate $(i, j)$ represents the normalised mean attention weight that agent $i$ assigns to services operated by agent $j$. To analyse the attention pattern, during the period in which the learnt mechanism provides the advantage of the greater sample efficiency, the weights are aggregated at training step 200K. Notably, the attention does not follow a uniform distribution, as there are weights concentrated along the main diagonal. This behaviour is consistent with the partial observability of the environment, since agents are only able to fully observe the services they control, so their own entities provide a richer and more informative signal. Nevertheless, attention is still assigned to competitor services, indicating that the visible public attributes of rival entities are also informative.

\subsubsection{How does decoupling representation learning from the actor affect the training dynamics?} \label{sec:detach_ablation}

The framework detaches the aggregated observation from the computational graph when computing the actor update, so that the GNN parameters are optimised only through the critics loss. The motivation behind this design choice is that, since they optimise different objectives, using both can lead to a more unstable training of the agents. To evaluate its impact, the default configuration is compared with a variant in which the representation module is jointly updated by the critic and the actor losses.

\begin{figure}[!ht]
    \centering
    \includegraphics[width=1\textwidth]{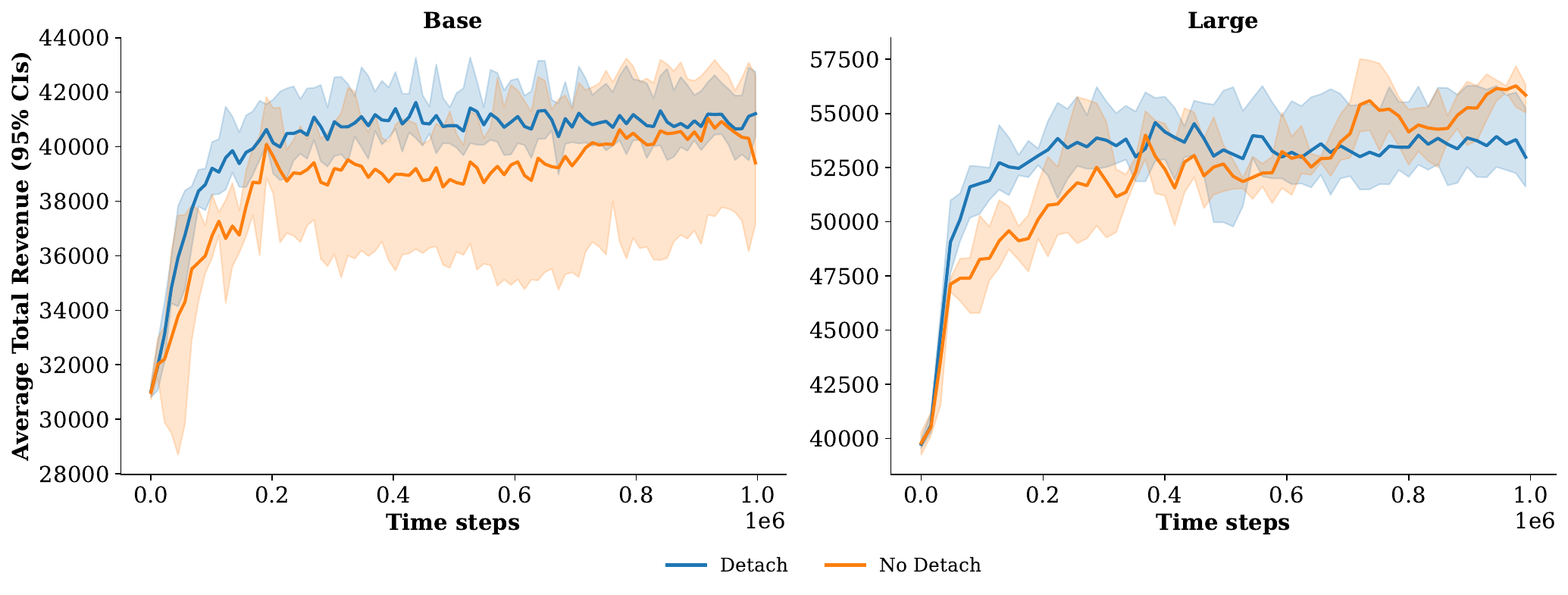}
    \caption{Ablation study comparing average total revenue of detach and no detach variants.}
    \label{fig:detach_ablation}
\end{figure}

Figure \ref{fig:detach_ablation} shows the average total revenue during training for both variants. In particular, in the first scenario, the detached configuration results in a policy with better performance and less variability across runs compared to the other approach, whereas in the second scenario, it is the no detach variant that achieves the higher revenue at the end of training. However, this figure only reflects the revenue obtained by the agents in total.

To quantify the stability difference at agent level, for each run, a stability metric is calculated as the standard deviation of the consecutive differences of the individual revenue (Table \ref{tab:detach_stability}). Intuitively, this metric is a discrete analogue of the variability of the learning curve derivative, so smaller values indicate that revenue increments from one step to the next are more consistent. More specifically, the detached variant achieves the lowest value for every agent in both scenarios, with this difference being more pronounced in the second case. This implies that decoupling representation learning from the actor stabilises the training dynamics of each agent individually, although there is a trade-off with a reduced total revenue.

\begin{table}[!ht]
    \caption{Training stability per agent measured as the standard deviation of consecutive differences of the individual revenue. The mean and the standard deviation were computed over three independent runs.}
    \small
    \begin{tabular*}{\textwidth}{@{\extracolsep\fill}lcccc}
        \toprule
        & \multicolumn{2}{c}{\textbf{Base}} & \multicolumn{2}{c}{\textbf{Large}} \\
        \cmidrule{2-5}
        & Detach & No Detach & Detach & No Detach \\
        \midrule
        Agent 1 & $\mathbf{455.7\pm70.7}$ & $498.9\pm144.0$ & $\mathbf{666.4\pm114.0}$ & $846.1\pm195.3$ \\
        Agent 2 & $\mathbf{363.9\pm215.5}$ & $441.2\pm13.6$ & $\mathbf{622.4\pm61.8}$ & $943.7\pm17.3$ \\
        Agent 3 & $\mathbf{521.7\pm39.2}$ & $537.1\pm100.9$ & $\mathbf{450.7\pm50.6}$ & $649.3\pm61.2$ \\
        Agent 4 & - & - & $\mathbf{356.2\pm37.7}$ & $535.9\pm106.1$ \\
        \bottomrule
    \end{tabular*}
    The lowest values are indicated in bold.
    \label{tab:detach_stability}
\end{table}

\subsubsection{What do the learnt graph embeddings capture about agent strategies?} \label{sec:embeddings}

To understand the information encoded by the learnt embeddings, t-SNE \cite{VanDerMaaten2008VisualizingT-SNE} visualisation was applied to the attention-weighted observations of the trained agents. Specifically, these visualisations project the aggregated observation vectors $\tilde{\mathbf{z}}^a$ that serve as input to each agent's actor and critic networks. For each scenario, the three experimental runs with different random seeds during the evaluation at the end of training are presented, with each point representing an agent's aggregated observation at a single time step. Each observation point is coloured in two ways: first, by its controlling agent, to reveal structural patterns in how different agents perceive the market, and second, by its associated revenue, to uncover strategic value distributions.

\begin{figure*}[!ht]
    \begin{center}
        \begin{subfigure}{0.9\textwidth}
            \includegraphics[width=1\linewidth]{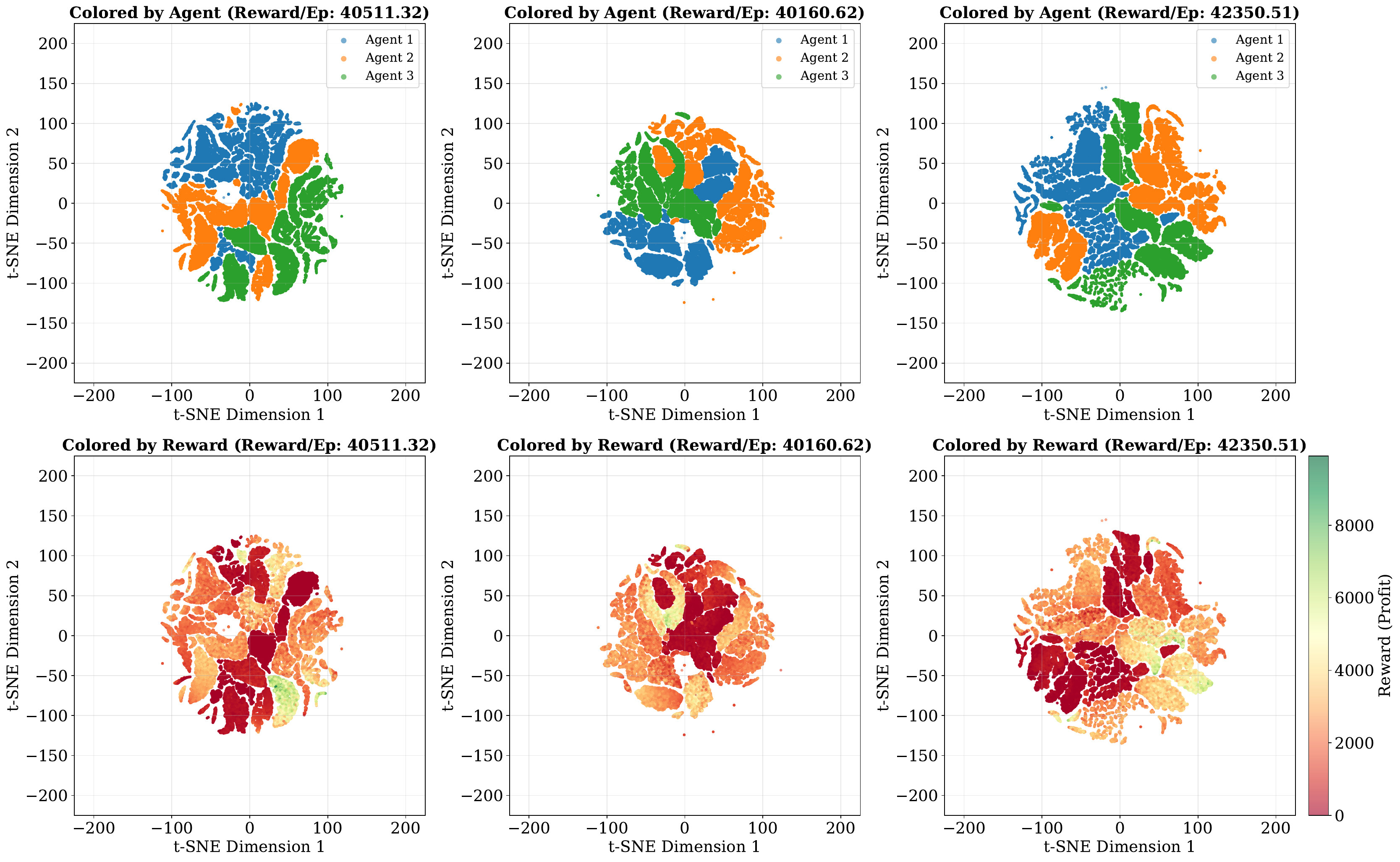}
            \caption{Base.}
        \end{subfigure}
        \begin{subfigure}{0.9\textwidth}
            \includegraphics[width=1\linewidth]{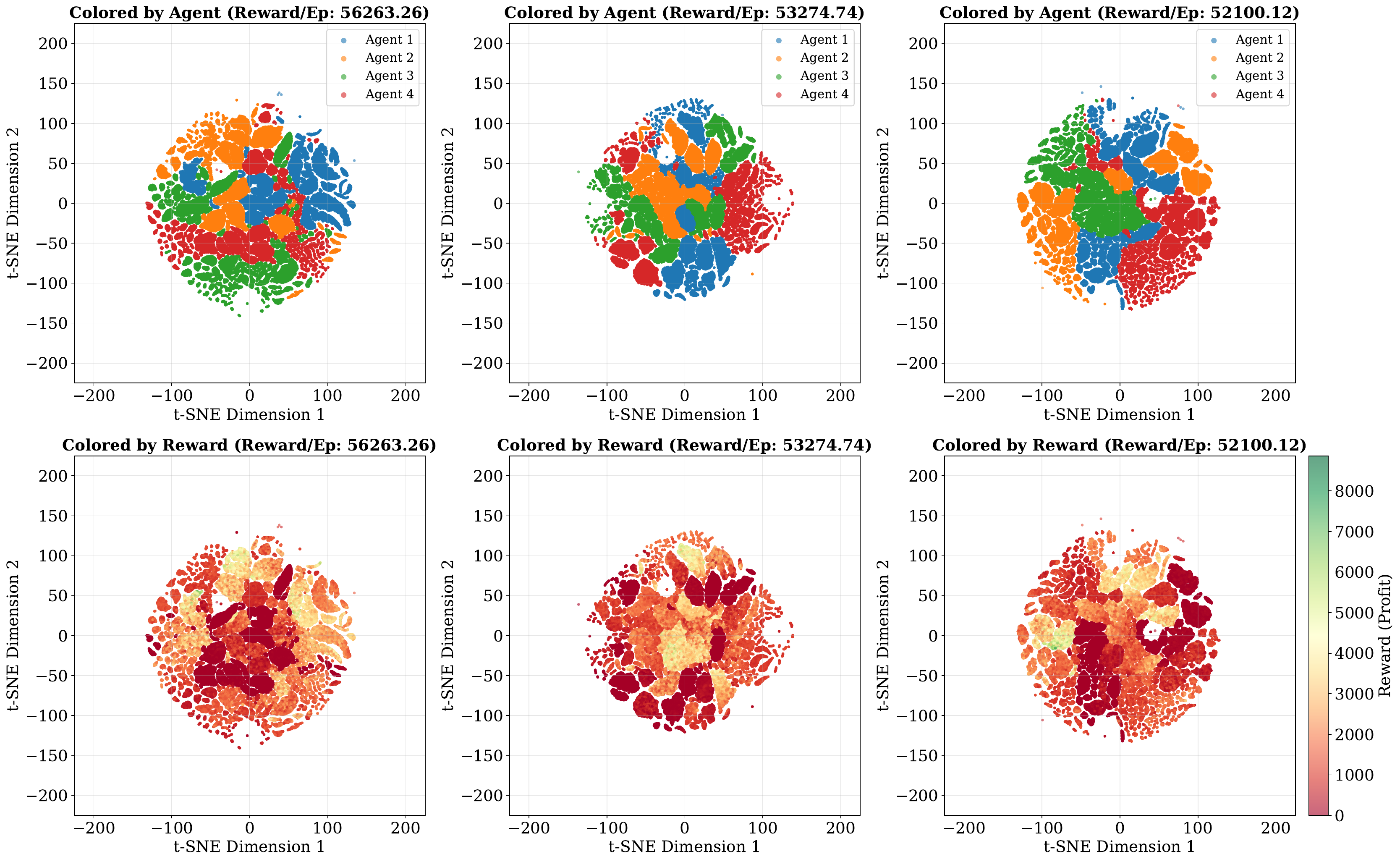}
            \caption{Large.}
        \end{subfigure}
    \end{center}
    \caption{t-SNE visualisation of learnt agent observations coloured by agent and reward. The clustering patterns suggest that the embeddings capture aspects of agent identity and market profitability.}
    \label{fig:tsne}
\end{figure*}

Figure \ref{fig:tsne} illustrates the projections of the aggregated observations for both scenarios. On the one hand, the agent-coloured projections show that the aggregated observations of different agents are mapped to distinct, contiguous regions of the latent space, with little overlap between them. This can be attributed to the fact that each agent learns its own representation, as each of them has its own GNN. Moreover, the partial observability of the environment that hides private ticket-sales data of competitor services encourages the same service to be encoded differently in each agent's view of the graph. On the other hand, the reward-coloured projections present a similar pattern with high and low revenue points clustered together, rather than being uniformly distributed. Therefore, this suggests that the learnt embeddings appear to encode who is acting and how profitable the market state is.

\section{Conclusions and future work}\label{sec:conclusions}

This work introduces an entity relational actor-critic framework for partially observable MARL with mixed competitive-cooperative settings. It encodes the environment as graph of operational units with heterogeneous edges, rather than decision-making agents or static infrastructure, connected by competition, coordination, and connectivity relations. Each agent learns a relational state representation through a multi-layer R-GCN, which is then aggregated into an agent-level observation via a learnt attention mechanism, and trains its policy under the CTDE paradigm by extending MATD3. The framework has been evaluated in a railway pricing reinforcement learning environment in two different scenarios of increasing market complexity, where it obtained a higher total revenue compared to the other MARL baselines considered. Moreover, it achieved this with smaller price increments and a higher percentage of passengers travelling compared to MATD3.

One potential benefit of the approach is the scalability with respect to the market structure, due to the compact fixed-size agent embeddings and the message passing paradigm that keep the model size independent of the number of services. This contrasts with the agent-based graph baseline GA-AC, in which each operator is encoded into a single node and whose effectiveness collapses in the larger scenario. According to the further studies, the competition and connectivity relations contribute the most to performance, while the coordination relation, which is the closest to the agent-based modelling, has the smallest impact. In addition, regarding the depth of the message passing, the two-layer configuration achieves the highest revenue, since it captures direct and connecting services with one transfer. Furthermore, the attention mechanism improves sample efficiency in the early stages of training, and detaching the representation module from the actor gradients has been shown to stabilise the individual agent training dynamics, even though it can reduce the total revenue.

As future work, it is suggested to explore methods where agents could directly discover which entities are connected to each other, rather than relying on predefined edge types. For instance, this could be achieved by using attention mechanisms that dynamically learn a relational structure based on observed state transitions and rewards. The framework would also benefit from modelling the temporal aspect within the graph structure itself. Currently, the entity graph is reconstructed at each time step based on static rules, but many real-world systems present relations that evolve over time. To this end, temporal graph networks \cite{Rossi2020} could be incorporated into the model to capture relational changes as the environment evolves. Finally, only off-policy methods are considered in this work, so future directions could extend the framework to on-policy algorithms in order to assess their potential advantages in this setting.

% --------------------
\section*{CRediT authorship contribution statement}
\textbf{Enrique Adrian Villarrubia-Martin:} Writing – original draft, Visualisation, Validation, Software, Methodology, Data curation, Conceptualisation. \textbf{David Muñoz-Valero:} Writing – review \& editing, Data curation, Software. \textbf{Luis Rodriguez-Benitez:} Writing – original draft, Resources, Methodology,  Formal analysis. \textbf{Giovanni Montana:} Writing – review \& editing, Supervision, Methodology, Conceptualisation, Formal analysis. \textbf{Luis Jimenez-Linares:} Supervision,  Project administration, Methodology, Conceptualisation, Funding acquisition.

% --------------------
\section*{Declaration of competing interest}
The authors declare that they have no known competing financial interests or personal relationships that could have appeared to influence the work reported in this paper.

% --------------------
\section*{Data availability}
Data will be made available upon request.

% --------------------
\section*{Acknowledgments}
This work was supported by grant PID2020-112967GB-C32 funded by MCIN/AEI/10.13039/501100011033 and by the ERDF A Way of Making Europe. It was completed when Enrique Adrian Villarrubia-Martin was a predoctoral fellow at the Universidad de Castilla-La Mancha funded by the European Social Fund Plus (ESF+) and in a visiting research stay at the University of Warwick funded by a mobility grant from the Universidad de Castilla-La Mancha for predoctoral students. Giovanni Montana acknowledges support from a UKRI AI Turing Acceleration Fellowship (EPSRC EP/V024868/1).

% --------------------
\appendix
\setcounter{table}{0}

\section{Hyperparameters} \label{sec:hyperparameters}

This section provides a comprehensive overview of the hyperparameters used to train the algorithms. For a standardised comparison, the default hyperparameters of each algorithm were used without additional tuning. Shared hyperparameters are presented in Table \ref{tab:shared_hyperparameters} and are used unless an algorithm-specific table assigns a different value. Algorithm-specific configurations are detailed in: Table \ref{tab:rache_hyperparameters} for RACHE, Table \ref{tab:maddpg_hyperparameters} for MADDPG, Table \ref{tab:matd3_hyperparameters} for MATD3, Table \ref{tab:maac_hyperparameters} for MAAC, and Table \ref{tab:gaac_hyperparameters} for GA-AC.

\newpage

\begin{table}[!ht]
    \caption{Shared hyperparameters between algorithms.}
    \small
    \centering
    \begin{threeparttable}
        \begin{tabular}{lc}
            \toprule
            Hyperparameter & Value \\
            \midrule
            Number of environments & 16 \\
            Optimiser & Adam \cite{Kingma2015Adam:Optimization} \\
            Buffer size & 1,000,000 \\
            Training steps & 1,000,000 \\
            Random policy episodes & 1,000 \\
            Evaluation episodes & 10,000 \\
            Number of hidden layers & 2 \\
            Hidden units per layer & 256 \\
            Non-linear activation function & ReLU \\
            Batch size & 1024 \\
            Discount factor ($\gamma$) & 0.99 \\
            Policy learning rate & 0.001 \\
            Critic learning rate & 0.001 \\
            Target smoothing coefficient ($\tau$) & 0.005 \\
            Steps per update & 100 \\
            Seeds & 0, 41, 73 \\
            Training environment seed & $\{\text{seed} + r \cdot 1000 \mid r \in \{0, 1, \ldots, 15 \}\}$ \\
            Evaluation environment seed & $\{ \text{seed} + r \cdot 100000 \mid r \in \{0, 1, \ldots, 15 \}\}$ \\
            Observation normalisation & Yes \\
            Reward scaling & 1000 \\
            \bottomrule
        \end{tabular}
    \end{threeparttable}
    \label{tab:shared_hyperparameters}
\end{table}

\begin{table}[!ht]
    \caption{RACHE hyperparameters.}
    \small
    \centering
    \begin{tabular}{lc}
        \toprule
        Hyperparameter & Value \\
        \midrule
        Categorical embedding dim & 4 \\
        Continuous embedding dim & 128 \\
        R-GCN learning rate & 0.001 \\
        R-GCN hidden dim & 128 \\
        Number of R-GCN layers & 2 \\
        Dropout & 0.1 \\
        Attention score function & MLP \\
        \bottomrule
    \end{tabular}
    \label{tab:rache_hyperparameters}
\end{table}

\begin{table}[!ht]
    \caption{MADDPG hyperparameters.}
    \small
    \centering
    \begin{threeparttable}
        \begin{tabular}{lc}
            \toprule
            Hyperparameter & Value \\
            \midrule
            Discount factor ($\gamma$) & 0.95 \\
            Policy learning rate & 0.01 \\
            Critic learning rate & 0.01 \\
            \multirow{2}{*}{Exploration noise} & Ornstein-Uhlenbeck \cite{Uhlenbeck1930OnMotion} \\
            & $\theta = 0.15$ $\sigma = 0.2$ \\
            \bottomrule
        \end{tabular}
    \end{threeparttable}
    \label{tab:maddpg_hyperparameters}
\end{table}

\begin{table}[!ht]
    \caption{MATD3 hyperparameters.}
    \small
    \centering
    \begin{tabular}{lc}
        \toprule
        Hyperparameter & Value \\
        \midrule
        Exploration noise & $\mathcal{N}(0, 0.1)$ \\
        Policy noise & $\mathcal{N}(0, 0.2)$ \\
        Policy noise clip & 0.5 \\
        Frequency delayed policy updates & 2 \\
        \bottomrule
    \end{tabular}
    \label{tab:matd3_hyperparameters}
\end{table}

\begin{table}[!ht]
    \caption{MAAC hyperparameters.}
    \small
    \centering
    \begin{tabular}{lc}
        \toprule
        Hyperparameter & Value \\
        \midrule
        Non-linear activation function & LeakyReLU \\
        Number of attention heads & 4 \\
        Number of updates per update cycle & 4 \\
        \bottomrule
    \end{tabular}
    \label{tab:maac_hyperparameters}
\end{table}

\begin{table}[!ht]
    \caption{GA-AC hyperparameters.}
    \small
    \centering
    \begin{tabular}{lc}
        \toprule
        Hyperparameter & Value \\
        \midrule
        Entropy regularisation coefficient ($\alpha$) & 0.2 \\
        \bottomrule
    \end{tabular}
    \label{tab:gaac_hyperparameters}
\end{table}

\bibliographystyle{elsarticle-num}
\bibliography{references}

\end{document}